\documentclass[pdflatex, referee, sn-apa]{sn-jnl}% APA Reference Style 
\usepackage{graphicx}%
\usepackage{multirow}%
\usepackage{amsmath,amssymb,amsfonts}%
\usepackage{amsthm}%
\usepackage{mathrsfs}%
\usepackage[title]{appendix}%
\usepackage{xcolor}%
\usepackage{textcomp}%
\usepackage{manyfoot}%
\usepackage{booktabs}%
\usepackage{algorithm}%
\usepackage{algorithmicx}%
\usepackage{algpseudocode}%
\usepackage{listings}%
\usepackage{subfig}
\usepackage{upquote}
\usepackage{pdflscape}
\usepackage{afterpage}
\usepackage{tabularx}
\usepackage{xltabular}
\usepackage{longtable}
\usepackage{hyperref}
\usepackage{comment}

\definecolor{codegreen}{rgb}{0,0.6,0}
\definecolor{codegray}{rgb}{0.5,0.5,0.5}
\definecolor{codepurple}{rgb}{0.58,0,0.82}
\definecolor{backcolour}{rgb}{0.95,0.95,0.92}

\lstdefinestyle{mystyle}{
    backgroundcolor=\color{backcolour},   
    commentstyle=\color{codegreen},
    keywordstyle=\color{magenta},
    numberstyle=\tiny\color{codegray},
    stringstyle=\color{codepurple},
    basicstyle=\ttfamily,
    breakatwhitespace=false,         
    breaklines=true,                 
    captionpos=a,                    
    keepspaces=true,                 
    numbersep=5pt,                  
    showspaces=false,                
    showstringspaces=false,
    showtabs=false,                  
    tabsize=4
}

\theoremstyle{thmstyleone}%
\theoremstyle{thmstyletwo}%
\usepackage{xspace}
\newcommand{\pymovements}{\texttt{pymovements}\xspace}
\newcommand{\review}{\textcolor{black}}
\theoremstyle{thmstylethree}%

\begin{document}

\title[Eye-Tracking-while-Reading Survey]{Eye-Tracking-while-Reading: A Living Survey of Datasets with Open Library Support}


\author*[1]{\fnm{Deborah N.} \sur{Jakobi}}\email{deborahnoemie.jakobi@uzh.ch}
\author[1, 2]{\fnm{David R.} \sur{Reich}}\email{davidrobert.reich@uzh.ch}
\author[2]{\fnm{Paul} \sur{Prasse}}\email{paul.prasse@uni-potsdam.de}
\author[1]{\fnm{Jana M.} \sur{Hofmann}}\email{janamara.hofmann@uzh.ch}
\author[1]{\fnm{Lena S.} \sur{Bolliger}}\email{lena.bolliger@uzh.ch}
\author[1, 3]{\fnm{Lena A.} \sur{J\"ager}}\email{lenaann.jaeger@uzh.ch}

\affil[1]{\orgdiv{Department of Computational Linguistics}, \orgname{University of Zurich}}
\affil[2]{\orgdiv{Department of Computer Science}, \orgname{University of Potsdam}}
\affil[3]{\orgdiv{Department of Informatics}, \orgname{University of Zurich}}

%\affil*[1]{\orgdiv{Department}, \orgname{Organization}, \orgaddress{\street{Street}, \city{City}, \postcode{100190}, \state{State}, \country{Country}}}


\abstract{Eye-tracking-while-reading corpora are a valuable resource for many different disciplines and use cases. Use cases range from studying the cognitive processes underlying reading to machine-learning-based applications, such as gaze-based assessments of reading comprehension. The past decades have seen an increase in the number and size of eye-tracking-while-reading datasets as well as increasing diversity with regard to the stimulus languages covered, the linguistic background of the participants, or accompanying psychometric or demographic data. The spread of data across different disciplines and the lack of data sharing standards across the communities lead to many existing datasets that cannot be easily reused due to a lack of interoperability. In this work, we aim at creating more transparency and clarity with regards to existing datasets and their features across different disciplines by i) presenting an extensive overview of existing datasets, ii) simplifying the sharing of newly created datasets by publishing a living overview online, \url{https://t.uzh.ch/1Yh}, presenting over 55 features for each dataset, and iii) integrating all publicly available datasets into the Python package \pymovements which offers an eye-tracking datasets library. By doing so, we aim to strengthen the FAIR principles in eye-tracking-while-reading research and promote good scientific practices, such as reproducing and replicating studies.}




\maketitle

\section{Introduction}\label{sec:intro}
For many people, reading is an everyday process and as such a necessary skill to master a variety of tasks in many areas of human life. The interaction of visual perception and language processing during reading opens a wide range of potential research topics. One of these is the investigation of cognitive processes that are involved in reading to understand how humans process language. Collecting and analyzing eye movements during reading has been shown to provide valuable insights into such processes, and therefore eye movements are considered a gold standard method in reading research \citep{Rayner1998, raynerCarroll2018, kliegl2006tracking}. The primary use case resulting from this tight connection of eye movements to human cognition is to develop psycholinguistic theories of reading \citep{just1980theory}. Even beyond this use case, eye movements are extremely valuable when it comes to, for example, investigating why certain individuals experience difficulties in reading in general, or why a given text might be hard to understand. The insights into human language processing can then be further used to, for example, investigate the cognitive plausibility of language models \citep{beinborn2023cognitive} or even enhance them using gaze data \citep[e.g.,][]{deng-etal-2024-fine}. The list of use cases of eye-tracking-while-reading data is long and keeps growing, as detailed in Section~\ref{sec:use-cases}. 

The number of use cases is expanded by the diversity of eye-tracking-while-reading data, which comes in many different forms. In general, participants read any type of reading material while their eye movements are being tracked.
The first very important characteristic of the data is the type of stimulus shown while collecting the data. Stimuli can range from isolated sentences to paragraphs or full-length texts, presented as individual sentences, entire paragraphs, multi- or single-page, and are typically categorized as either controlled or naturalistic. Controlled, minimal pair stimuli are hand-crafted sentences,  carefully designed to test specific linguistic hypotheses under tightly constrained conditions. In psycholinguistics, such controlled settings are commonly used to test well-defined theoretical predictions. Typically, the experiment focuses on
an individual phenomenon or the interaction of a very limited number of phenomena, such as investigating garden path sentences~\citep{Pickering1998gardenpath}, locality or surprisal effects~\citep{DEMBERG2008193}, or syntactic word order manipulations of any kind \citep{rayner2013ProcessingCanonicalWord, traxler2002ProcessingSubjectObject}.
In contrast, naturalistic corpora involve reading real-world texts such as news articles, narratives, or scientific content, allowing researchers to observe a broader and more representative range of linguistic phenomena. 
\review{These corpora are particularly useful for investigating a range of research questions, such as studying passage-level comprehension~\citep{prasse2024cogstate}, or broad-coverage evaluation of psycholinguistic theories of sentence  processing \citep[e.g.,][]{engelmann2019EffectProminenceCue, gibson_dependency_2000, futrell2020LossyContextSurprisalInformationTheoretic}.}

However, the division between controlled and naturalistic data is not always clear-cut. Some corpora use semi-naturalistic stimuli, including hand-crafted, hand-edited, or hand-selected sentences. These aim to balance experimental control with ecological validity, allowing researchers to target specific linguistic features\footnote{Typically, these are linguistic constructions that are theoretically relevant, but relatively infrequent in naturalistic texts.} while maintaining a higher degree of naturalness.

The type of text used within naturalistic corpora also influences the range of possible analyses. For instance, reading domain-specific scientific texts may be suitable for studying expert reading strategies, while tabloid texts may offer insights into everyday reading behavior. 
Another distinction concerns the type of task or the instructions that the reader receives. Readers can be asked to perform tasks during reading, such as annotation or summarization tasks, they might have to answer questions concerning the text at some point through the experiment, or perform a repeated reading of an already known text. Similarly, the reading instructions may differ, for example, by requiring the reader to read aloud, silently, or within a time limit. Yet another aspect of eye-tracking-while-reading datasets is the availability of labels or annotation that is published together with the data. For example behavioral data recorded for each participant that may be directly related to the eye-tracking experiment, or not (such as psychometric tests), or annotations of the stimulus such as lexical frequency, surprisal, and many more.

 The growing number of available eye-tracking datasets, along with recent expansions in participant size, linguistic diversity, text types, and research domains~\citep{ reich2025EyeTrackingNLP}, has fueled increasing interest in eye-tracking-while-reading data. This makes it more attractive to reuse existing datasets, while at the same time making it harder to keep track of them. When reusing data, however, the specific characteristics of a dataset are crucial, as they directly determine its suitability for a given research objective---be it hypothesis-driven analyses or exploratory studies with more naturalistic stimuli. A clear overview of available datasets is therefore essential to support informed research decisions.

However, there are challenges regarding the publication and sharing of the data. No standardized format exists for sharing the data, nor one platform where all datasets can be found, but rather discipline-specific platforms. Many different, often unconnected disciplines work with eye-tracking data and dataset creators typically use sharing platforms common within their discipline. There are also no standards for what metadata to publish. Certain metadata is crucial for further processing, such as the eye-to-screen distance, which is required to calculate gaze velocity. However, there have been recent advances in this area to push such standards \citep{dunn2023MinimalReportingGuideline, JakobiKrakowczyk2024reporting}, and by means of this present work, we aim to further establish these standards by creating a comprehensive overview of existing datasets with the objective of achieving greater transparency. 

The goal of this paper is therefore threefold. First, we provide a comprehensive review of existing (semi-)naturalistic eye-tracking-while-reading datasets including information on the participants, languages, materials, as well as  descriptive statistics on the stimulus texts and the gaze recordings. Most importantly, we outline the corpora's available data formats, and the meta-data released. 

Second, we extend the impact of the overview presented in this paper by publishing an \textit{extensive, filterable, online table} that lists the existing datasets including all information on them that is openly available. By doing so, we aim at making the overview easier to use and, most importantly, editable (e.g., it is possible to add so far unpublished information on already listed datasets as well as adding more datasets to the overview). The table is found at the following link: \href{https://t.uzh.ch/1Yh}{https://t.uzh.ch/1Yh}. The overview can easily be extended and adjusted by dataset authors themselves, allowing researchers to make their datasets easily findable (see Section~\ref{sec:add}).

Third, a standalone contribution is to simplify the reuse of existing datasets even further by integrating all publicly available datasets presented in this review into our open-source Python package \pymovements~\citep{Krakowczyk_pymovementsETRA2023}. \pymovements is a Python library for preprocessing and analyzing eye-tracking data that also comes with an \texttt{R} API\footnote{\url{https://pymovements.readthedocs.io/en/stable/tutorials/R-tutorial.html}}. 
The package includes an interface for downloading and processing publicly available datasets \citep{pm-datasets-2025}. Including a dataset into \pymovements means that all available data formats can be downloaded through its API. For example, if a given dataset is published as raw data (x-, and y-screen coordinates including timestamp), those can be downloaded and further processed using \pymovements. If the dataset authors have published their data including both raw data and fixation data, the entire dataset can be downloaded using the \pymovements interface. This feature allows the re-user of the respective dataset to either use the preprocessed data provided by the original authors, or to perform their own preprocessing using a wide range of preprocessing functionalities provided by \pymovements, or their own methods (refer to Section~\ref{sec:pm} for more details on how to work with \pymovements). The possibility of easily applying the same preprocessing to all datasets eventually leads to more standardized, transparent, and, most importantly, reproducible preprocessing pipelines.   

By providing an extensive overview over existing eye-tracking-while-reading corpora as well as integrating available datasets into \pymovements, our aim is to encourage the publication of datasets to follow the FAIR principles (Findable, Accessible, Interoperable, Reusable;~\citealp{wilkinson2016FAIRGuidingPrinciples}), and the publication of the data, metadata and quality reports as outlined by \citet{JakobiKrakowczyk2024reporting}. We foster a publication standard in which considerations about the  format and metadata of the dataset go beyond the original use case or research question for which the data were collected, but recognize the utility of a completely reproducible data pipeline for other researchers and use cases. 

In sum, the contribution of this  work is to i) present a comprehensive overview of (semi-)naturalistic eye-tracking-while-reading datasets, ii) publish a filterable table online to easily access the dataset overview, and iii) integrate the datasets into \pymovements for easy reuse. By doing the aforementioned, we contribute to strengthen the FAIR principles as we hope to encourage future dataset authors to publish their data including all data formats, metadata and all information as presented in the table.

\section{Use cases of eye-tracking-while-reading data}
\label{sec:use-cases}
Eye movements in reading are influenced by many different factors. Most importantly, eye movements reflect the underlying cognitive processes involved in reading, which makes them a powerful tool for psycholinguistic research. Eye-tracking has therefore been widely used to study language comprehension at the lexical, morphological, syntactic, semantic, and pragmatic levels~\citep{inhoff2019RegressionsReading, Rayner1998}. In this context, statistical models are typically employed to investigate how linguistic variables impact gaze patterns.
Eye-tracking also provides insights into individual differences in reading, since eye movements are highly idiosyncratic \citep{kuperman2011indiff, haller2026ReplicateMeIf, MakowskiECML2018}. This can include research on specific populations, such as dyslexic readers~\citep{pavlidis1985dyslexia, benfattoScreeningDyslexiaUsing2016}, or individuals with autism spectrum disorder~\citep{Howard2017}. Another well-studied domain is reading comprehension, where researchers examine how higher-level comprehension, beyond the sentence level, is reflected in eye movements \citep{meziere2023UsingEyetrackingMeasures}. Another area of psycholinguistic eye-tracking-while-reading research is the generation of eye movements using computational cognitive models such as SWIFT \citep{engbert2005SWIFTDynamicalModel}, E-Z-reader \citep{reichle2003EZReaderModel}, \"Uber-Reader \citep{veldre_towards_2020}, or SEAM \citep{rabe2024SEAMIntegratedActivationcoupled}. The models implement different theories of reading to generate human scanpaths.

In recent years, the scope of eye-tracking-while-reading has expanded beyond traditional psycholinguistics. In addition to statistical modeling, machine learning methods have been increasingly applied to eye movement data for reader inference---that is, predicting reader characteristics from their gaze behavior. Reader-level data, such as scores obtained from psychometric tests, or native language, can serve as labels in classification tasks. For instance, given recordings from dyslexic and non-dyslexic children, machine learning models can learn to classify each scanpath as being generated by a dyslexic or non-dyslexic reader. Such approaches have been used to infer native language~\citep{berzak-etal-2017-predicting, ReichETRA2022}, language proficiency~\citep{berzak-etal-2018-assessing}, dyslexia~\citep{bjornsdottir-etal-2023-dyslexia, haller-etal-2022-eye,raatikainen2021detection, vajs2023accessible}, or reading comprehension~\citep{copeland2014readingcomp, shubi-etal-2024-fine, ReichETRA2022, ahn2020PredictingReadingComprehension}. EyeBench is a recently launched project that presents a benchmark for reader inference from eye movements during reading, illustrating that eye movements are informative across a wide range of tasks \citep{shubiEyeBenchPredictiveModeling}.

Eye movement data has also entered NLP research, where it serves two broad purposes. First, gaze is used to augment language models or models used for downstream applications. Because eye movements unfold over time and their data structure hence differs fundamentally from the static nature of text, integration into text-based models is non-trivial. \review{One approach is to construct gaze embeddings for each word in the text. Such embeddings can be obtained by, for example, encoding a scanpath using neural models. This allows for extracting vectors representing the eye movements on a given text. These representations can then be combined with representations of each word individually, by, for example, adding or concatenating the two representations \citep{lopez-cardona2025seeing}.} Alternatively, gaze can inform attention mechanisms (e.g., assigning lower attention weights to words with shorter total fixation duration;~\citealp{sood2023improvingattention}), or the input text can be reordered according to a scanpath such that the words follow the chronological order in which they were fixated by a given reader \citep{deng2023-augmented, deng-etal-2024-fine, yang2023plm}.
Such gaze-augmented models have been shown to improve performance in a wide range of NLP tasks, including sentiment analysis \citep{yang2023plm, mishra2017LeveragingCognitiveFeatures, barrett-etal-2018-sequence, ren-xiong-2021-cogalign}, named entity recognition \citep{hollenstein-zhang-2019-entity, ren-xiong-2021-cogalign}, paraphrase generation and sentence compression \citep{sood2020improving, klerke-etal-2016-improving}, relation extraction \citep{ren-xiong-2021-cogalign}, NLU tasks (Natural Language Understanding, GLUE benchmark; \citealp{deng2023-augmented, deng-etal-2024-fine}), readability assessment \citep{gonzalez-garduno-sogaard-2017-readability}, dependency parsing \citep{strzyz-etal-2019-towards}, question answering \citep{malmaud-etal-2020-bridging}, or grammatical acceptability classification \citep{bondar_colagaze_2025}.

Second, NLP methods are evaluated using eye movement data. This is possible because, as has been introduced before, there is a tight connection between human language processing and eye movements. This connection can be leveraged, for example, to automatically assess the difficulty of a text generated by a language model. Instead of letting a human assign a score, eye movements can be tracked while reading the text and evaluated. Reading measures such as the rate of regressions can then be used to assess the readability of the text \citep{klein2025eyetrackingbasedcognitive}. NLP applications for machine translation~\citep{Doherty2010, sajjad-etal-2016-eyes}, or summarization can be evaluated in a similar fashion~\citep{ikhwantri-etal-2024-analyzing}. 

Finally, eye-tracking-while-reading corpora are used to train data-driven, machine-learning-based generative models of eye movements in reading \citep{deng2023eyettention, deng2026EyettentionIIDualSequence, bolliger-etal-2023-scandl, bolliger2025Scandl2}. Synthetically generated eye movements can be used for pretraining models for reader inference or overcome the lack of eye movement data at application time. Generative models typically require large amounts of data to perform well, and the recent advances in deep neural generative modeling of eye movements has only been possible due to the large increase of available data.

This listing of use cases and studies is not exhaustive; it serves the purpose of underlining the utility of eye-tracking-while-reading corpora across disciplines. 
It is important to realize that many of these use cases, be it classification tasks or the analysis of specific reader characteristics, can be addressed using the same dataset. There is no need to collect new data for each and every use case, the only requirement being that data which possesses relevant characteristics has been collected and published in a FAIR way \citep{wilkinson2016FAIRGuidingPrinciples}.

\section{Eye-tracking-while-reading data formats}
\label{sec:data-formats}

 \begin{figure}
     \centering
     \includegraphics[width=0.9\linewidth]{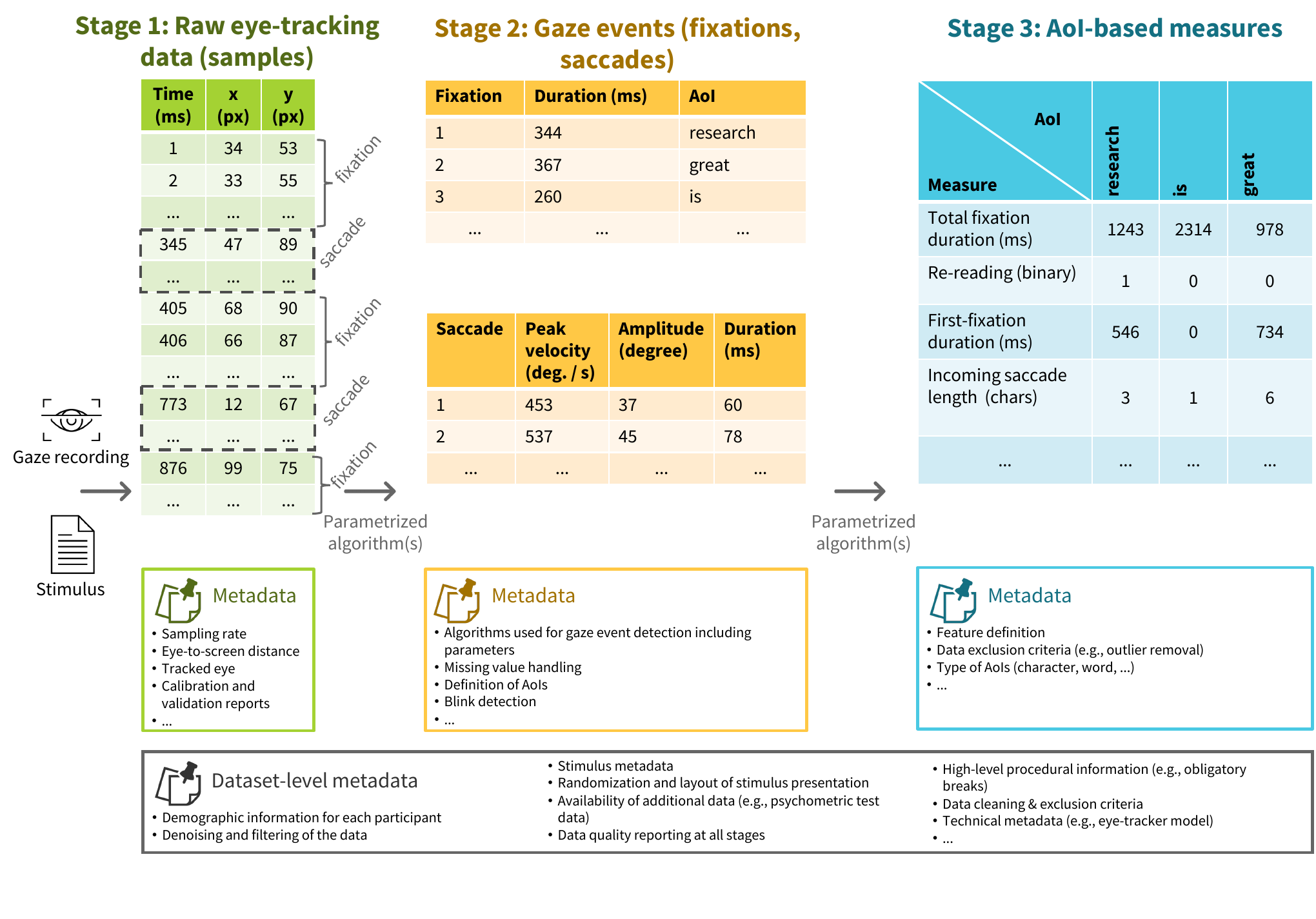}
     \caption{Preprocessing stages for eye-tracking-while-reading data. Stage 1: Raw data consisting of x- and y-screen coordinates or yaw and pitch degrees of visual angle, and a timestamp. Stage 2: Gaze event data (fixations and saccades). Stage 3: AoI-based measures (area-of-interest). Each stage requires certain metadata and produces new metadata. The data from one stage is converted to the next stage by parametrized algorithms. If the algorithms and their parameters are unknown, the pipeline cannot be fully reproduced.}
     \label{fig:stages}
 \end{figure}
 
Eye-tracking-while-reading data undergoes several preprocessing steps, each of which produces different data formats corresponding to different stages of processing. In addition, metadata is needed for each stage, or defined or generated at a stage. An overview of the preprocessing stages and their metadata is provided in Figure~\ref{fig:stages}. As different use cases and analyses require the data at a different stage, information about how the data has been preprocessed, and ideally making the data available at all stages is crucial when publishing eye-tracking-while-reading data.
The initial input for preprocessing the eye-tracking data is the gaze recording and the stimulus shown on the screen. 

The \textit{first stage} is the raw data (time series data), consisting of x- and y-coordinates in pixels on the screen, and a timestamp, depending on the sampling frequency. The format of the raw data which is written by the eye-tracking device differs greatly between different eye trackers and it often requires to parse the gaze recording to extract the samples and relevant metadata (e.g., display resolution).   
In some cases, the data recording generated by the eye tracker must first be converted into a human readable format. For certain eye trackers, this introduces an additional ``stage zero'', namely the conversion of the original file created at the time of recording into a gaze recording which can be parsed to extract the samples. 
The following stages are based on this sample-level data. Publishing only the raw data is consequently the minimal requirement to be able to use the data for all use cases.

The \textit{second stage} aggregates the samples into gaze events (discrete, chronological sequence). For each gaze event, a set of features can be calculated, such as the duration of the event or the mean location of all event samples in pixels on the screen (for more examples see columns for Stage 2 in Figure~\ref{fig:stages}). During reading, the eyes alternate between two events: fixations---when the gaze rests on a location on the screen and visual input is obtained---and saccades, which are rapid ballistic movements between fixation points during which visual intake is minimal. While these gaze events are embedded in the raw data, they must be detected by grouping consecutive samples into events. A variety of algorithms exist for this purpose, each making specific assumptions about which features can be used to detect an event type. For example, the widely used I-DT (dispersion-threshold identification) algorithm assumes that all samples belonging to a fixation lie within a spatial window of a given size; the dispersion threshold defining this window can be adjusted. A group of samples that are too dispersed will not be classified as a single fixation \citep{salvucci2000fixations}. Other algorithms detect events based on velocity which is motivated by the fact that saccades are fast movements while fixations are more stationary, and consequently the gaze velocity is much lower during a fixation \citep[e.g,][]{engbert2003MicrosaccadesUncoverOrientation}. Depending on the algorithm used and the thresholds chosen, the on- and offsets of the detected events can vary greatly \citep{reich2025EvaluatingGazeEvent}.

The \textit{third stage} consists of area-of-interest (AoI)-based measures (discrete, non-chronological sequence). In addition to the gaze recording, this stage needs information on the stimulus shown while the gaze was recorded. Typically, the stimulus is defined in terms of areas in pixel on the screen which correspond to different units of the stimulus. These units are often words, but can be multiple words or sub-words such as morphemes. The definition of the AoIs depends on the use case, and consequently does the definition of the AoI-based measures.
Examples for AoI-based measures are word-level measures used in reading research, such as the total fixation duration on a word or the presence of an outgoing regression. This stage aggregates the data even further, summarizing the gaze events which can be linked to stimulus units relevant for the analysis. 

Note that the stimulus can be mapped to the gaze data at each of the three stages.  If a single sample falls inside one of the stimulus unit areas, one sample can be mapped to the respective stimulus unit. For fixations, the mean position can be mapped to the stimulus in a similar fashion resulting in scanpaths. Even saccades can be mapped to the stimulus by, for example, determining the start and end positions and mapping these to the respective stimulus unit.

Each state requires and produces different metadata, which also includes parameters that need to be chosen for the preprocessing algorithms. For example, in order to calculate fixations based on velocity, the gaze velocity needs to be calculated first, which requires knowledge about the screen dimensions and eye-to-screen distance. Figure~\ref{fig:stages} lists important metadata for each stage as well as metadata at the dataset-level.

While these high-level descriptions of each stage seem to result in comparable datasets at each stage, the data at each step can differ greatly, even if the same algorithm is applied \citep{reich2025EvaluatingGazeEvent}. Different implementations and different parametrization cause these differences. Assuming a minimal fixation duration of 50\,ms compared to 150\,ms changes how fixations are detected which in consequence influences, for example, AoI-based metrics.
 It is therefore essential for data re-users to know which preprocessing algorithms were applied and how they were parameterized. Publicly providing this information ensures reproducibility, allows the impact of different methodological choices to be assessed, and ultimately contributes to improving preprocessing practices across research domains. Additionally, each preprocessing step includes cleaning the data and potentially the removal of data points considered irrelevant for the current purpose. All preprocessing decisions may impact all analyses conducted with the data \citep{ landes2025ImpactPreprocessingClassification}.

\section{A brief history of eye-tracking-while-reading corpora}

\begin{figure*}[ht]
    \begin{center}
\centering
	  \subfloat[\label{fig:cum_datasets}Number of datasets by data modality.]{\includegraphics[trim= 0 0 0 0,clip,width=0.49\textwidth,keepaspectratio]{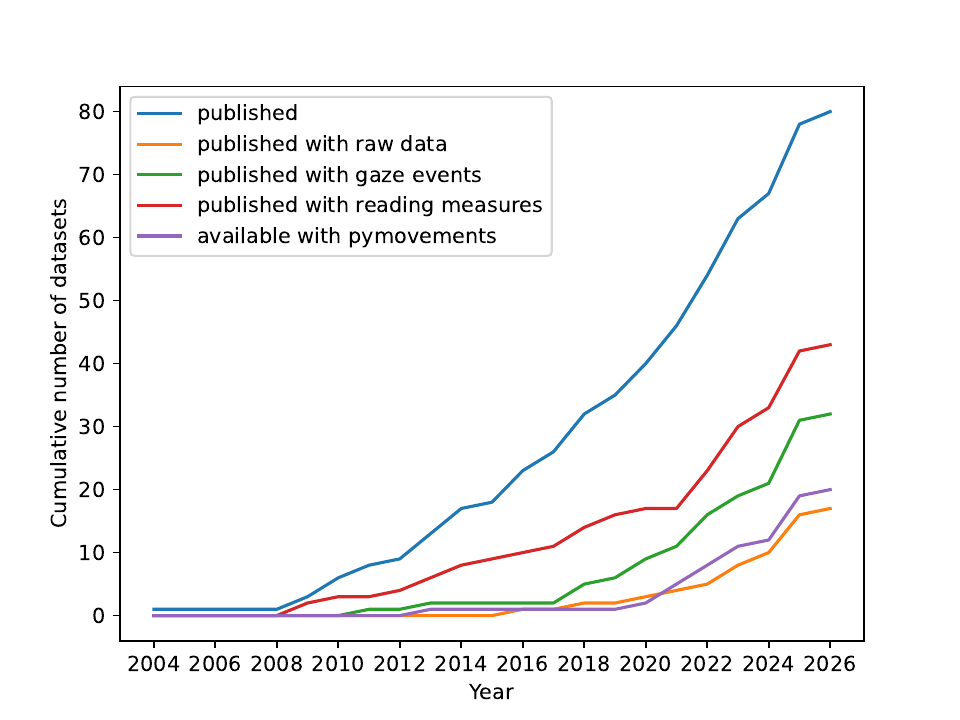}}
	  \subfloat[\label{fig:cum_languages} Number of stimulus languages.% / language families.
      ]{\includegraphics[trim= 0 0 0 0,clip,width=0.49\textwidth,keepaspectratio]{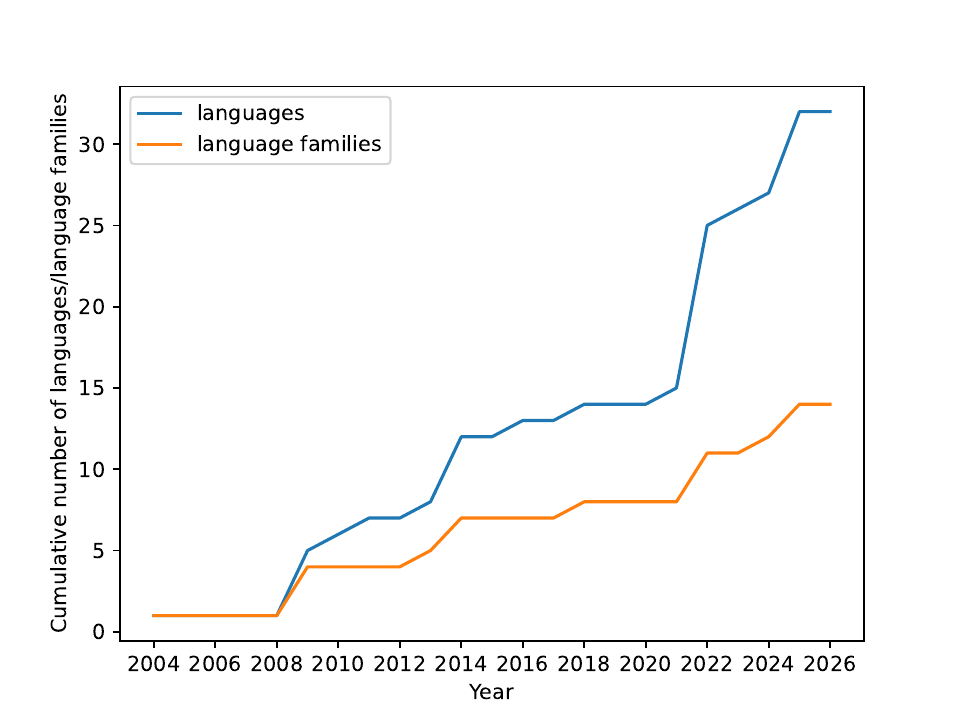}}
\end{center} 
    \caption{Cumulative overview of eye-tracking-while-reading datasets over time. (a) Cumulative number of published datasets by data modality, indicating which datasets are supported by the open-source library pymovements \citep{Krakowczyk_pymovementsETRA2023}. (b) Cumulative number of stimulus languages and language families\protect\footnotemark represented in published datasets. In both sub-figures, the x-axis denotes the year of publication.}
    \label{fig:cum_plots}
\end{figure*}

In recent years, the number of eye-tracking-while-reading datasets has grown substantially \citep{reich2025EyeTrackingNLP}. This increase is evident not only in the sheer number of datasets but also in their scope: newer datasets often include larger participant samples \citep[e.g.,][]{celer2022, Siegelman2022mecol1wave1, siegelman2025mecol1wave2}, cover a broader range of reading tasks \citep[e.g.,][]{berzak2025OneStop360ParticipantEnglish, Hollenstein2018zuco1}, encompass additional languages \citep[e.g.,][]{kuperman2025mecol2wave2, kuperman-2023-mecol2-wave1}, and involve more diverse participant populations \citep[e.g.,][]{hollenstein-etal-2022-copenhagen, yaneve2016asd}. For comparison, one of the earliest datasets, Dundee \citep{dundee}, published in 2003, included 20 participants reading English or French texts. While many datasets still remain below 100 participants, recent examples such as CELER \citep{celer2022} involve several hundred participants, and the MECO L1 datasets even exceed one thousand readers, albeit on different languages  \citep{Siegelman2022mecol1wave1, siegelman2025mecol1wave2}. Please see Figure~\ref{fig:cum_languages} for a chronological overview of stimulus languages and language families represented in the included datasets.

\footnotetext{The language familyies for each language are chosen as follows: Basque: Basque, Brazilian Portuguese: Romance, Cantonese (trad. chars.): Sinitic, Chinese: Sinitic, Czech: Slavic, Danish: Germanic, Dutch: Germanic, English: Germanic, Estonian: Finno-Ugric, Finnish: Finno-Ugric, French: Romance, German: Germanic, Greek: Hellenic, Hebrew: Semitic, Hindi: Indo-Aryan, Icelandic: Germanic, Italian: Romance, Japanese: Japonic, Korean: Koreanic, Mandarin: Sinitic, Mandarin (simp. chars.): Sinitic, Mandarin (trad. chars.): Sinitic, Mongolian: Mongolian, Norwegian: Germanic, Persian: Iranian, Portugese: Romance, Russian: Slavic, Serbian: Slavic, Spanish: Romance, Swedish: Germanic, Turkish: Turkic, Uighur: Turkic, and Urdu: Indo-Aryan. Please refer to our \href{https://t.uzh.ch/1Yh}{online table} for a list of all language families represented in Figure~\ref{fig:cum_languages}.}

Another very notable development is the trend towards more openly available datasets including their metadata. While datasets used to be made available only upon request or not at all, it has become more common to publish the data through established open data channels like OSF\footnote{\url{https://osf.io/}} or GitHub\footnote{\url{https://github.com/}}. This trend is also reflected in the publication of datasets that are published including all available metadata, scripts and the complete preprocessing pipeline \citep[e.g,][]{berzak2025OneStop360ParticipantEnglish, jakobi2025PoTeCGermanNaturalistic, frank2024EyetrackingwithEEGCoregistrationCorpus, hollenstein-etal-2022-copenhagen, bolliger_emtec_2025}. This development is reflected in Figure~\ref{fig:cum_datasets}, which shows that datasets are increasingly published in different formats. A very early initiative to share eye-tracking-while-reading data more openly and completely is the \href{https://gitup.uni-potsdam.de/pmr2}{Potsdam Mind Research Repository (PMR2)}. The repository has very recently been revived. It contains dedicated repositories for different studies, including eye-tracking data and scripts. The first version of this \href{https://web.archive.org/web/20220124153837/http://read.psych.uni-potsdam.de/}{repository} has been one of the first to present an overview of a lab's different eye-tracking studies and their data with the specific goal being to encourage researchers to reuse the data, replicate findings, and reproduce pipelines. In addition, users were explicitly encouraged to add new datasets or new analysis of already published datasets.

Datasets have also become richer in metadata, with more detailed documentation of eye-tracking hardware, experimental setups, and additional measures such as psychometric test results (e.g., IQ or working memory capacity) and comprehension questions that extend beyond basic attention checks and allow investigation of comprehension at different levels \citep{berzak2025OneStop360ParticipantEnglish, Jakobi2025MultipleyeETRA}.

The trend of creating larger and more diverse datasets has been impressively demonstrated by the MECO multi-lab data collection initiative \citep[The Multilingual Eye-tracking Corpus;][]{Siegelman2022mecol1wave1,siegelman2025mecol1wave2, kuperman-2023-mecol2-wave1, kuperman2025mecol2wave2}. Its datasets include large participant samples split into L1 and L2 reading groups, collected across multiple data collection sites, spread over more than 40 countries, and including more than 30 languages.\footnote{\url{https://meco-read.com/}} Such collaborations require careful coordination and standardization, but they enable large-scale data collections by distributing the workload. A similar multi-lab effort is MultiplEYE \citep{Jakobi2025MultipleyeETRA}, which explicitly promotes open research data (ORD) standards and provides an open-source, transparent preprocessing pipeline designed for long-term reuse.

\section{Eye-tracking-while-reading corpus review}
\label{sec:review}
This section provides an overview of the different eye-tracking-while-reading datasets. The section is further sub-divided according to the language of the stimuli. All datasets are presented in tables which list the name of the dataset as well the most important features (e.g., number of participants or number of stimulus items). As not all datasets report the same features or use different implementations of the same feature, such as the number of words per item \textit{or} overall, the features in each table may differ across languages. An overview of all features\footnote{Please note that the table itself as well as its features can be extended. The online table is therefore the most up-to-date version of the datasets as well as the features.} that have been considered and are presented in the online overview is presented in Table~\ref{tab:column_descriptions}.

In order to increase the readability of the tables presented in this paper, only a selected number of features is reported. A more detailed description of each dataset is published in conjunction with this review and can be found at the following link: \href{https://t.uzh.ch/1Yh}{https://t.uzh.ch/1Yh}. More than 55 features are reported overall, although not all of them are available for each dataset. The online table is curated by the authors of this paper to ensure consistent presentation. However, dataset authors can and are explicitly encouraged to request changes and add new datasets through our website\footnote{\url{https://t.uzh.ch/1Yv}}. We strongly encourage dataset authors to inform us about additional datasets to be added to the online table, or information about existing datasets which is not yet included in the table. Please refer to Section~\ref{sec:add} for more information on how to add information or datasets to our overview.

\begin{footnotesize}
\centering
\begin{xltabular}{\textwidth}{>{\raggedright\hsize=.3\hsize}X|>{\raggedright\hsize=.7\hsize}X}\caption{The features that are reported for each dataset in the living online table including their description. Please note that the same descriptions are also added to the online table and that not all features are available for each dataset.} \label{tab:column_descriptions}\\
\hline
\textbf{Column Name} & \textbf{Description} \\
\midrule
\endfirsthead
\multicolumn{2}{l}%
{\tablename\ \thetable{} -- continued from previous page} \\
\toprule
\textbf{Column Name} & \textbf{Description} \\
 \\\midrule
\endhead
\multicolumn{2}{c}{General information} \\\hline
Name & The name of the dataset \\
\hline
Data accessibility & If and how the eye-tracking data is accessible. Restricted means that the data can be accessed but only by, e.g., writing an e-mail to the authors, signing up, or filling in a form. \\
\hline
pymovements & If the dataset is integrated into the Python package \href{https://pymovements.readthedocs.io/en/stable/datasets/index.html}{\pymovements'} dataset library. \\
\hline
Full name & The full name of the dataset if the name is an abbreviation. \\
\hline
Dataset characteristics & Any characteristics of the dataset that could make it particularly interesting or special. \\
\hline
\multicolumn{2}{c}{Participants} \\\hline
\# Participants & Number of participants. \\
\hline
Age range & The age range min-max of all participants. \\
\hline
Age mean$\pm$SD & The mean and standard deviation of the age across all participants. \\
\hline
Native language & The native languages of the participants if known. \\
\hline
Inclusion criteria & Criteria applied to include participants, if known. \\
\hline
Exclusion criteria & Criteria applied to exclude participants, if known. \\
\hline
Other characteristics & Other notable characteristics of the participants. \\
\hline
Reading instructions & The instructions participants received before reading the text. \\
\hline
\multicolumn{2}{c}{Stimuli} \\\hline
\# Items & Number of items overall. \\
\hline
\# Items per subject & Number of items each participant read. \\
\hline
Mean$\pm$SD \# words per item & Mean and standard deviation of words per item. \\
\hline
Mean$\pm$SD \# chars per item & The mean and standard deviation of the number of characters per item. \\
\hline
Mean$\pm$SD \# chars per word & The mean and standard deviation of the number of characters per word. \\
\hline
Mean$\pm$SD \# chars seen per participant & The mean and standard deviation of the number of characters seen by each participant. \\
\hline
Mean$\pm$SD \# words seen per participant & The mean and standard deviation of the number of words seen by each participant. \\
\hline
Total \# chars all items & The total number of characters across all items. \\
\hline
Total \# words all items & The total number of words across all items. \\
\hline
Stimulus description & A short textual description of the stimuli. \\
\hline
Stimulus language & The language of the stimulus. \\
\hline
Stimulus language family & The language family of the stimulus texts. The language family indicated for one language is chosen one or two levels up when illustrating the families as a tree such that multiple languages are summarized with that term but the category is not too broad. For example, Indo-European languages are categorized as Romance, Germanic, etc., to avoid the very broad category of Indo-European. Purely geographical categories (e.g., Western, Eastern) were skipped and the higher-ranking category was chosen. \\
\hline
Stimulus naturalness & Degree of naturalness of the stimuli. Naturalistic: existing texts, only adapted when necessary, for example to make them self-contained. Partially constructed: created by combining elements, such as by selecting target words or by controlling certain characteristics across all stimuli but choosing the stimuli from existing sources. Constructed: created specifically for the experiment, while still aiming to be naturalistic. \\
\hline
Stimulus length category & Broad categories for the length of the texts, e.g., sentences or passages. \\
\hline
Stimulus source & The link of the stimulus source.  \\
\hline
Comprehension questions & Whether there were comprehension questions. \\
\hline
Text annotation & Whether the stimulus texts are annotated or not and the type of annotation. \\
\hline
Stimulus license & How the stimulus texts are licensed.  \\
\hline
\multicolumn{2}{c}{Available data} \\\hline
Raw data & Whether raw data is made available for this dataset. \\
\hline
Gaze events & Whether fixation data is made available for this dataset. \\
\hline
Reading measures & Whether reading measures are made available for this dataset. \\
\hline
Scripts & Whether any programming scripts are made available for this dataset. Can include scripts for analysis, experiment presentation, or preprocessing. \\
\hline
Data license & The license the data is shared under. \\
\hline
Notes on availability & Whether there is a special procedure, login, form etc. required to access the data. \\
\hline
\multicolumn{2}{c}{Lab setup} \\\hline
Eye-tracker & The brand and type of eye-tracker used. \\
\hline
Mount & Eye-tracker mount and type of head support used for the participants. \\
\hline
Sampling frequency (Hz) &  The sampling frequency that has been used in the experiment to collect the data. \\
\hline
Eye-to-screen distance (cm) &  The distance in centimeters measures from the participants eyes to the monitor in a 90 degrees angle. \\
\hline
Eye-to-camera distance (cm) &  The distance in centimeters measures from the participants eyes to the camera. \\
\hline
Monitor & The model and possibly a description of the monitor used to present the stimuli. \\
\hline
Monitor size (cm) & The size of the monitor in cm used to present the stimuli (width and height). \\
\hline
Resolution & Resolution of the screen in pixels. \\
\hline
\multicolumn{2}{c}{Stimulus layout} \\\hline
Text presentation & The format and the way the text is presented on the screen. \\
\hline
Font & The font name. \\
\hline
Font size & The size of the font with the respective unit (if known). \\
\hline
Monospaced & Whether the font was monospaced or not. \\
\hline
Character per visual angle & How many characters per visual angle were displayed. \\
\hline
Font color & The color of the font the stimuli were presented in. \\
\hline
Spacing & The line spacing if there are multiple lines on each page. \\
\hline
Background color & The color of the background behind the stimuli. \\
\hline
\multicolumn{2}{c}{Preprocessing} \\\hline
Preprocessing tool & The tool that has been used to preprocess the eye-tracking data. Can be a (licensed) software, (public) code base or any other tool.  \\
\hline
Fixation detection algorithm and parameters & The name of the algorithm which was used to detect fixations and any parameters or thresholds that were chosen. \\
\hline
Vertical fixation correction & Indicates whether the fixation locations have been post-hoc corrected vertically.  \\
\hline
Notes on preprocessing & Any other relevant information on the preprocessing. \\
\hline
\end{xltabular}%
\end{footnotesize}

\subsection{Multilingual}
\label{sec:multilingual}
\review{The largest multi-lingual eye-tracking-while-reading corpus to date is the Multilingual Eye-Movements Corpus (MECO), which, in total, covers 22 languages and includes 1189 native speakers reading the same 12 encyclopedic texts (requiring no academic background) in their respective language \citep{Siegelman2022mecol1wave1, siegelman2025mecol1wave2}. The data collection has been split into two waves, as presented in Table~\ref{tab:multilingual}. MECO-L2 has also been collected using only English materials and non-native participants which is included in Section~\ref{sec:english}}.
%The largest multi-lingual eye-tracking-while-reading corpus to date is the Multilingual Eye-Movements Corpus (MECO), which comprises data from more than 1000 participants in total across 13 languages split into two data collection waves. Participants read 12 encyclopedic texts in their native language, with topics selected to avoid the need for an academic background. In the first wave, data from 535 participants has been collected \citep{Siegelman2022mecol1wave1}. In a second wave, data from another 654 participants has been collected \citep{siegelman2025mecol1wave2}. The 14 stimulus languages are partially overlapping with the first wave while adding some new languages (MECO-L2 has also been collected using only English materials and non-native participants which is included in Section~\ref{sec:english}).
Another corpus is WebQAmGaze \citep{ribeiro2023webqamgaze}, which includes data from 19 participants reading both long and short naturalistic texts in German, English, and Spanish. As the data has been collected with a webcam, it has lower temporal and spatial precision compared to high-precision eye trackers.
The Ghent Eye-tracking Corpus, GECO, includes 33 participants (both Dutch-English bilinguals and English monolinguals) reading an entire novel in English and Dutch \citep{Cop2016geco}. Its companion dataset, GECO-CN, contains eye-tracking data from 30 Chinese participants reading the same novel in English and Chinese \citep{Sui2022gecocn}.
The Dundee corpus contains data from 20 participants reading naturalistic English and French texts \citep{dundee}.
The Potsdam-Allahabad Hindi Eyetracking Corpus comprises data from 30 participants reading 153 naturalistic single sentences in Hindi and Urdu \citep{Husain-Vasishth-Srinivasan-2014-hindi}.
For the Little Prince Corpus, 60 participants read the story \textit{The Little Prince} by Antoine de Saint-Exupéry either in Mandarin (simplified Chinese characters) or Cantonese (traditional Chinese characters) with 30 participants for each language \citep{li-etal-2023-little-prince}.
For the Developmental Reading corpus \citep{feng2009TimeCourseHazard, feng2009OrthographyDevelopmentReading}, 159 participants with either English or Chinese as their native language were recruited from different age groups (children and adults). They were asked to read different stories which were partially comparable between the languages.
The Grammatical/Ungrammatical dataset contains data from 98 native German or English speakers reading 92 grammatical and ungrammatical sentences in the respective native language \citep{vasishth2010ShorttermForgettingSentence}.
 Two related datasets exist, presenting 120 and 86 sentences with differing morphological structures to 48 and 30 participants, respectively \citep{yan2014EyeMovementsGuided}. The participants read half of the sentences in Uighur and the other half in Mandarin (simplified characters).
The Story Reading dataset \citep{feng2009TimeCourseHazard} contains data from 42 participants reading novels in their native language (English, Chinese, Japanese, and Korean).
For more detailed  information on these multilingual datasets, see Table~\ref{tab:multilingual}.

\begin{landscape}
\begin{footnotesize}
\begin{xltabular}{1.6\textwidth}
{>{\raggedright\hsize=.1\hsize}X|>{\raggedright\hsize=.05\hsize}X>{\raggedright\hsize=.08\hsize}X>{\raggedright\hsize=.09\hsize}X>{\raggedright\hsize=.07\hsize}X>{\raggedright\hsize=.09\hsize}X>{\raggedright\hsize=.1\hsize}X>{\raggedright\hsize=.08\hsize}X>{\raggedright\hsize=.06\hsize}X>{\raggedright\hsize=.1\hsize}X>{\raggedright\hsize=.08\hsize}X}\caption{Datasets with multilingual stimuli. For a description of the columns please refer to Table~\ref{tab:column_descriptions}.} \label{tab:multilingual} \\
\toprule
Name & \# Participants & Age range & Age mean±SD & \# Items & Mean±SD words per item & Stimulus language & Stimulus naturalness & Stimulus length category & Eye-tracker & Sampling frequency (Hz) \\
\\\midrule
\endfirsthead
\multicolumn{11}{c}%
{\tablename\ \thetable{} -- continued from previous page} \\
\toprule
Name & \# Participants & Age range & Age mean±SD & \# Items & Mean±SD words per item & Stimulus language & Stimulus naturalness & Stimulus length category & Eye-tracker & Sampling frequency (Hz) \\
 \\\midrule
\endhead
Developmental Reading & 159 &  &  & 16 & 225.5 (English), 149.5 (Chinese) & English, Chinese  & Naturalistic & Text passages & EyeLink I & 250 \\
Dundee & 20 &  &  &  &  & English, French & Naturalistic & Text passages & Dr Bouis Oculometer Eyetracker & 1000 \\
GECO & 33 &  & Bilinguals: 21.2±2.2; monolinguals: 21.8±5.6 & 1 &  & English, Dutch & Naturalistic & Text passages & EyeLink 1000 & 1000 \\
GECO-CN & 30 & 20-29 & 25.3±2.6 & 1 &  & English, Mandarin (simp. chars.) & Naturalistic & Text passages & EyeLink 1000 Plus & 1000 \\
Grammatical / Ungrammatical & 98 &  &  & 92 &  & German, English & Constructed & Single sentences & EyeLink I (English), IView-X (German)  & 250 (English), 240 (German) \\
MECO L1 $1^{st}$ Wave & 535 & 18-45 &  & 12 &  & German, Dutch, English, Greek, Hebrew, Italian, Russian, Spanish, Turkish, Korean, Norwegian, Finnish, Estonian & Naturalistic & Text passages & EyeLink Portable Duo, EyeLink 1000, EyeLink 1000 Plus & 1000 \\
MECO L1 $2^{nd}$ Wave & 654 & 18-58 &  & 12 &  & Basque, Brazilian Portuguese, Mandarin (simp. chars.), Mandarin (trad. chars.), Danish, English, German, Hindi, Icelandic, Norwegian, Russian, Serbian, Spanish, Turkish & Naturalistic & Text passages & EyeLink Portable Duo, EyeLink II, EyeLink 1000, EyeLink 1000 Plus  & 1000; Serbian: 500 \\
Morphological Structure I & 48 &  &  & 120 & 9.2±1.2 & Uighur, Mandarin (simp. chars.) &  &  & EyeLink II & 500 \\
Morphological Structure II & 30 &  &  & 86 & 7.3±1.6 & Uighur, Mandarin (simp. chars.) & Partially constructed & Single sentences & EyeLink II & 500 \\
Potsdam-Allahabad Hindi Eyetracking Corpus & 30 &  &  & 153 &  & Hindi, Urdu & Naturalistic & Single sentences & SMI iView X HED & 500 \\
Story Reading & 42 &  &  &  &  & English, Chinese, Japanese, Korean & Naturalistic & Text passages & EyeLink I (Korean study), EyeLink II (all other studies) & 250 (EyeLink I), 500 (EyeLink II) \\
The Little Prince Corpus & 60 &  & Cantonese: 21.7, Mandarin: 28.8 &  &  & Mandarin (simp. chars.), Cantonese (trad. chars.) & Naturalistic & Text passages &  &  \\
WebQAmGaze & 194 &  &  & 38 &  & German, English, Spanish & Naturalistic & Text passages & Webcam & 24.93 (mean) \\
\bottomrule
\end{xltabular}%
\end{footnotesize}
\end{landscape}%

\subsection{Basque}
The MECO L1 $2^{nd}$ wave contains a Basque subset as specified in Section \ref{sec:multilingual}.

\subsection{Chinese}
% all chinese languages are grouped here, i.e. Mandarin and Cantonese, and the script used
This section contains all datasets with stimuli in a Chinese language. This includes Cantonese and Mandarin as well as the different scripts (traditional or simplified Chinese characters). 
The Hong Kong Corpus of Chinese Sentence and Passage Reading \citep{Wu2023HKC} includes 96 participants reading 300 single-line sentences and 7 multi-line passages in Mandarin (simplified characters). 
\citet{Zhang2022chinese}  collected data from 1,718 participants reading in total 7,577 different Mandarin sentences (simplified characters; Eye-movement Measures on Words in Chinese Reading). 
\cite{Zang2018zh-word-len} collected a Mandarin (simplified characters) dataset used to study Chinese word length effects based on data from 30 participants reading 90 constructed sentences that were judged for their naturalness.
The BSC (Beijing Sentence Corpus) focuses on 150 single sentences chosen from the People's Daily in Mandarin (simplified characters; strong political tones have been removed) read by 60 participants \citep{Pan2021BSC}, with the BSC II providing the same stimuli in traditional characters, again for 60 participants \citep{Yan2025}. A third version of the BSC contains 48 participants reading sentences from the original BSC but with manipulations of the font size \citep{shu2011FontSizeModulates}.
Further, the ADEGBTS corpus (A Dataset for Exploring Gaze Behaviors in Text Summarization) includes data from 50 participants reading 100 naturalistic text passages and summarizing each passage after reading it \citep{yi2020gaze-sum}.
The Reading Attention dataset contains  29 participants reading 60 Mandarin (simplified characters) text passages each associated with a specific query \citep{li2018reading-attention}. The participants were asked to judge the relevance of the passage for the query.
\citet{yan2019EyeMovementControl} collected eye movement data in Mandarin (simplified characters) of 84 participants reading 200 naturalistic sentences (Eye Movement Control in Chinese Reading).
See Table~\ref{tab:chinese} for more information.
 
Several multilingual datasets contain Chinese subsets: GECO-CN, MECO L1 $2^{nd}$ wave (Mandarin in both simplified and traditional characters), the Story Reading corpus, the Developmental Reading corpus, Morphological Structure I \& II, as well as The Little Prince Corpus which is a parallel corpus of Mandarin (simplified characters) and Cantonese (traditional characters). Please refer to Section~\ref{sec:multilingual} for more information.

\begin{landscape}
\begin{footnotesize}
\begin{xltabular}{1.6\textwidth}
{>{\raggedright\hsize=.1\hsize}X|>{\raggedright\hsize=.025\hsize}X>{\raggedright\hsize=.1\hsize}X>{\raggedright\hsize=.1\hsize}X>{\raggedright\hsize=.05\hsize}X>{\raggedright\hsize=.1\hsize}X>{\raggedright\hsize=.1\hsize}X>{\raggedright\hsize=.1\hsize}X>{\raggedright\hsize=.1\hsize}X>{\raggedright\hsize=.1\hsize}X>{\raggedright\hsize=.1\hsize}X>{\raggedright\hsize=.1\hsize}X}\caption{Datasets with Chinese stimuli. For a description of the columns please refer to Table~\ref{tab:column_descriptions}.}\label{tab:chinese} \\
\toprule
Name & \# Participants & Age range & Age mean±SD & \# Items & Mean±SD words per item & Mean±SD chars per item & Stimulus language & Stimulus naturalness & Stimulus length category & Eye-tracker & Sampling frequency (Hz) \\
\midrule
\endfirsthead
\multicolumn{12}{c}%
{\tablename\ \thetable{} -- continued from previous page} \\
\toprule
Name & \# Participants & Age range & Age mean±SD & \# Items & Mean±SD words per item & Mean±SD chars per item & Stimulus language & Stimulus naturalness & Stimulus length category & Eye-tracker & Sampling frequency (Hz) \\
 \midrule
\endhead
ADEGBTS & 50 &  & 23.1±1.1 & 100 &  & 502 & Mandarin (simp. chars.) & Naturalistic & Text passages & Tobii EyeTracking 4C & 100 \\
BSC & 60 &  & 22.0±2.6 & 150 &  &  & Mandarin (simp. chars.) & Naturalistic & Single sentences & EyeLink II & 500 \\
BSC II & 70 &  &  & 150 & 11.2±1.6 & 21±2.5 & Mandarin (trad. chars.) & Naturalistic & Single sentences &  &  \\
BSC font size manipulation & 48 &  &  & 120 & 9.7±1 & 17.9±1.2 & Mandarin & Naturalistic & Single sentences & EyeLink II & 500 \\
Chinese Reading & 1718 &  &  & 7577 &  & 22.48 & Mandarin (simp. chars.) & Naturalistic & Single sentences & EyeLink 1000 & 1000 \\
Chinese Word Length Effect & 30 &  & 24.0±2.0 & 90 &  & 19.0±2.0 & Mandarin (simp. chars.) & Constructed & Single sentences & EyeLink 1000 Plus &  \\
Eye Movement Control in Chinese Reading & 84 &  &  & 200 & 14.7±1.3 & 8.1±1.0 & Mandarin (simp. chars.) & Naturalistic & Single sentences & EyeLink    & 1000 \\
Hong Kong Corpus & 96 &  & 26.0±3.64 & 307 &  &  & Mandarin (simp. chars.) & Naturalistic & Both single sentences and text passages & EyeLink 1000 & 1000 \\
Reading Attention & 29 & 17-28 &  & 60 &  &  & Mandarin (simp. chars.) & Naturalistic & Text passages & Tobii X2-30 &  \\
\bottomrule
\end{xltabular}
\end{footnotesize}
\end{landscape}%

\subsection{Czech}
ETDD70, Eye-Tracking Dataset for Classification of Dyslexia, is a Czech dataset \citep{dostalova2025ETDD70EyeTrackingDyslexia}. The dataset includes 70 dyslexic and non-dyslexic children reading multiple sentences aloud. See Table~\ref{tab:czech} for more detailed information on Czech datasets.

\begin{footnotesize}
\begin{xltabular}{\textwidth}
{>{\raggedright\hsize=.1\hsize}X|>{\raggedright\hsize=.12\hsize}X>{\raggedright\hsize=.15\hsize}X>{\raggedright\hsize=.16\hsize}X>{\raggedright\hsize=.17\hsize}X>{\raggedright\hsize=.15\hsize}X>{\raggedright\hsize=.15\hsize}X}\caption{Dataset with Czech stimuli. For a description of the columns please refer to Table~\ref{tab:column_descriptions}.} \label{tab:czech} \\
\toprule
Name & \# Participants & Age range & Stimulus naturalness & Stimulus length category & Eye-tracker & Sampling frequency (Hz)  \\
\midrule
ETDD70 & 70 & 9-10 & Partially constructed & Text passages & SMI RED 250 remote & 250\\
\bottomrule
\end{xltabular}
\end{footnotesize}

\subsection{Danish} 
CopCo, the Copenhagen Corpus, contains data from 57 participants who read 20 texts \citep{hollenstein-etal-2022-copenhagen}. 25 participants are native Danish speakers, 19 are dyslexic native speakers, and 13 are non-native speakers of Danish. See Table~\ref{tab:danish} for more information. 
Moreover, the second wave MECO L1 $2^{nd}$ wave contains a Danish subset, please refer to Section~\ref{sec:multilingual}.

%\begin{landscape}
\begin{footnotesize}
\begin{xltabular}{\textwidth}
{>{\raggedright\hsize=.1\hsize}X|>{\raggedright\hsize=.15\hsize}X>{\raggedright\hsize=.15\hsize}X>{\raggedright\hsize=.1\hsize}X>{\raggedright\hsize=.15\hsize}X>{\raggedright\hsize=.15\hsize}X>{\raggedright\hsize=.1\hsize}X>{\raggedright\hsize=.1\hsize}X}\caption{ For a description of the columns please refer to Table~\ref{tab:column_descriptions}.} \label{tab:danish} \\
\toprule
Name & \# Participants & Age range & \# Items & Stimulus naturalness & Stimulus length category & Eye-tracker & Sampling frequency (Hz) \\\midrule
\endfirsthead
\multicolumn{8}{c}%
{\tablename\ \thetable{} -- continued from previous page} \\
\toprule
Name & \# Participants & Age range & \# Items & Stimulus naturalness & Stimulus length category & Eye-tracker & Sampling frequency (Hz) \\\midrule
\endhead
CopCo & 57 & 21-62 & 20 & Naturalistic & Text passages & EyeLink 1000 Plus & 1000 \\
\bottomrule
\end{xltabular}
\end{footnotesize}
%\end{landscape}%

\subsection{Dutch}
The RaCCooNS (Radboud Coregistration Corpus of Narrative Sentences) dataset includes eye-tracking and EEG (electroencephalography) co-registration data from 37 participants reading 200 naturalistic narrative Dutch sentences, enabling multimodal analyses of sentence processing \citep{frank2024EyetrackingwithEEGCoregistrationCorpus}. The Mental Simulation Corpus contains readings of three literary short stories by 102 participants, paired with comprehension questions aimed at assessing participants' mental simulations during reading \citep{mak2019mentalsimulation}. DEMONIC (Dutch Eye-Movements ONline Internet Corpus) and DMORPH both feature single-line Dutch sentences. DEMONIC includes 55 participants who read 224 sentences per subject \citep{kupermanEffectWordPosition2010demonic}. 
The DMORPH stimulus sentences include 156 Dutch sentences including bi-morphemic target words with 19 different derivational suffixes (and 136 filler sentences), offering fine-grained insights into morphological processing \citep{kupermanProcessingTradeoffsReading2010demorph}. DMORPH includes data from 28 participants.
See Table~\ref{tab:dutch} for more information.
Moreover, GECO and MECO L1 $1^{st}$ wave both include a Dutch subset as specified in Section \ref{sec:multilingual}.

%\begin{landscape}
\begin{footnotesize}
\begin{xltabular}{\textwidth}
{>{\raggedright\hsize=.15\hsize}X|>{\raggedright\hsize=.09\hsize}X>{\raggedright\hsize=.1\hsize}X>{\raggedright\hsize=.08\hsize}X>{\raggedright\hsize=.1\hsize}X>{\raggedright\hsize=.13\hsize}X>{\raggedright\hsize=.15\hsize}X>{\raggedright\hsize=.1\hsize}X>{\raggedright\hsize=.1\hsize}X}\caption{Datasets with Dutch stimuli. For a description of the columns please refer to Table~\ref{tab:column_descriptions}.} \label{tab:dutch} \\
\toprule
Name & \# Participants & Age mean & \# Items & Mean±SD words per item & Stimulus naturalness & Stimulus length category & Eye-tracker & Sampling frequency (Hz) \\
\midrule
\endfirsthead
\multicolumn{9}{c}%
{\tablename\ \thetable{} -- continued from previous page} \\
\toprule
Name & \# Participants & Age mean & \# Items & Mean±SD words per item & Stimulus naturalness & Stimulus length category & Eye-tracker & Sampling frequency (Hz) \\
\midrule
\endhead
DEMONIC & 55 &  &  & 10.9±2.7 & Naturalistic & Single sentences & EyeLink II & 500 \\
DMORPH & 28 &  & 292 & 11.6±2.2 & Naturalistic & Single sentences & EyeLink II & 500 \\
Mental Simulation Corpus & 102 & 23 & 3 &  & Naturalistic & Text passages & EyeLink 1000 & 1000 \\
RaCCooNS & 37 & 26.2 & 200 &  & Naturalistic & Single sentences & EyeLink 1000 & 1000 \\
\bottomrule
\end{xltabular}
\end{footnotesize}
%\end{landscape}%

\subsection{English}
\label{sec:english}
% fully en
A wide range of eye-tracking-while-reading corpora exist for English texts, spanning diverse participant profiles and stimulus types.
The largest among them is the MECO L2 dataset, which includes data from 543 participants with various native languages reading 12 English texts \citep{kuperman-2023-mecol2-wave1}. In a second wave, more MECO L2 data has been collected adding an additional 660 participants with different language backgrounds reading the same 12 texts \citep{kuperman2025mecol2wave2}.
Similarly large-scale is CELER (365 participants; Corpus of Eye Movements in L1 and L2 English Reading) with data from 69 native and 296 non-native speakers reading 156 English sentences from the Wall Street Journal. The non-native speakers have five different native languages (Japanese, Arabic, Spanish, Portuguese and Chinese) and half of the sentences in the stimulus corpus are uniquely read by one participant \citep{celer2022}.
For the GazeBase dataset reading task, a total of 322 participants read up to 18 passages of a poem over a three year period \citep{Griffith2021gazebase}. In the GazeBase-VR dataset, data from 407 participants was collected. The participants performed one reading task among five different tasks whilst wearing an eye-tracking enabled virtual reality head set \citep{lohrGazeBaseVRLargescaleLongitudinal2023}.
The OneStop Eye Movements corpus (360 participants) includes 30 aligned texts at varying reading levels \citep{berzak2025OneStop360ParticipantEnglish}.
The Provo Corpus contains data from 84 participants reading 55 passages \citep{Luke2017provo}.
SB-SAT includes 95 participants reading texts from the Scholastic Assessment Test (SAT; \citealp{ahn2020PredictingReadingComprehension}). 
For the UCL Corpus, 205 sentences were sampled from English novels, and eye-tracking data is available for 43 participants \citep{frankReadingTimeData2013ucl}. 

Several CFILT (Computation for Indian Language Technology) datasets are available for English. For the CFILT sarcasm dataset, 7 non-native speakers of English read 1,000 sentences of which 350 had been previously labeled as sarcastic and 650 as non-sarcastic \citep{Mishra-Kanojia-Bhattacharyya-2016}. Participants had to rate the sentences as either positive or negative while their eye movements were tracked. 
A similar dataset is the CFILT sentiment complexity dataset where 5 participants were asked to rate the sentiment (positive, negative, or objective) of 1,059 English sentences  \citep{joshi-etal-2014-measuring}.
The CFILT scanpaths dataset contains data from 16 participants (including 3 linguistic experts) reading 32 paragraphs from (simple) Wikipedia articles. The participants annotated the texts for the effort to read them such that scanpath complexity can be studied \citep{mishra2017scanpath}. 
The CFILT text quality dataset includes data from 20 fluent English speakers reading 30 texts from different sources which were asked to rate the text quality given three properties (organization, coherence, and cohesion; \citealp{mathias-etal-2018-eyes}). 
The CFILT essay grading dataset includes 8 fluent English speakers reading and grading 48 essays while their eye movements were being tracked \citep{mathias-etal-2020-happy}.
The last dataset of this group is the CFILT coreference dataset where 14 participants read 22 texts while annotating the texts for coreferences \citep{cheri-etal-2016-coreference}.

Other corpora include the ASD Data with 109 participants of which 56 are diagnosed with ASD (Autism Spectrum Disorder) while the others are part of a control group \citep{yaneve2016asd}. All participants read text passages from different sources. 
A corpus constructed to study parafoveal vision by placing target words inside or outside of the parafoveal view contains 48 participants reading 40 text passages \citep[][Text Passages corpus]{parker2017passage-reading}.
MQA-RC is a reading comprehension dataset comprising 28 participants reading 32 movie plots and answering corresponding comprehension questions \citep{sood2020interpreting} taken from the benchmark MovieQA dataset \citep{tapaswi_movieqa_2016}.
The TECO dataset \citep{nahatame_teco_2024} includes eye movements from a total of 41 Japanese participants, who learned English as a second language, while reading 30 English text passages from the Eiken test.

The Reading Brain Project aims to investigate how readers of different ages and reading abilities comprehend scientific texts \citep{hsuNeurocognitiveSignaturesNaturalistic2019,
liReadingComprehensionL12019,
follmerWhatPredictsAdult2018}. To date, two datasets have been collected, with participants performing a reading task whilst undergoing simultaneous eye-tracking and fMRI (Functional Magnetic Resonance Imaging) scanning. 51 English native speakers are requested to read five short texts about a scientific topic. In the The Reading Brain Project L2 Dataset, 56 native speakers of Mandarin were tasked with reading the same five English texts. The participants were divided into two groups: one residing in the US and the other one in China during the data collection period.
IITB-HGC \citep{maharaj_eyes_2023} comprises the data of five participants assessing the correctness of a claim given its context. The 500 context-claim pairs were taken from the FactCC benchmark dataset, used to detect hallucination in automatically generated text \citep{kryscinski_evaluating_2019}.
50 Norwegian participants read 40 English text passages, with idioms used literally and figuratively in the Not Batting an Eye Dataset \citep{kyriacou_not_2024}. 
To assess the effects of tasks on human reading, \citet {hahn_modeling_2023} collected data from 20 Native English speakers \review{reading 20 news articles under two task conditions. The tasks included a preview condition, where participants first read a question on the text, then they read the text, and then they saw the question again with four answer choices and had to select one answer. In the no-preview condition, the question was not presented at the beginning of the trial, but only after the text had been read.}
EMTeC (Eye Movements on Machine-Generated Texts Corpus) contains eye‑tracking data from 107 native English speakers who read 42 machine‑generated one‑page texts \citep{bolliger_emtec_2025}.
CoLAGaze \citep{bondar_colagaze_2025} provides eye‑tracking for 42 native English speakers who read 306 sentences (153 grammatical–ungrammatical pairs) sourced from CoLA \citep{warstadt2019neural}. 
For GGTG (Gaze-guided text generation) 18 English texts had been generated using a Language Model which applied a mechanism to control the text generation by gaze. The eye-tracking dataset then contains data from 24 participants, including both native and non-native speakers of English, reading the generated texts \citep{sauberli2026ControllingReadingEase}. 
OASST-ETC comprises eye‑tracking data from 24 expert raters who evaluated LLM responses to OASST1 prompts (each participant inspected subsets of prompts and up to two candidate responses) \citep{lopez-cardona2025OASSTETCDatasetAlignment}.
InteRead contains eye‑tracking from 50 participants who read a 28‑page fiction excerpt with experimentally triggered interruptions \citep{zermiani_interead_2024}.
The Alzheimer dataset contains data from 162 older adults (79 memory clinic patients including participants diagnosed with Alzheimer's disease, mild cognitive impairment, or subjective memory complaints and 83 participants in a control group) who read a standardized paragraph (International Reading Speed Texts) and completed additional tasks \citep{jang_classification_2021}.
The Multi Choice Answering study recorded eye movements from 71 examinees (mostly non‑native English speakers) while they solved two controlled multiple‑choice comprehension items \citep{corbaci2022LatentGrowthModeling}.
Finally, the ZuCo corpora combine eye-tracking with EEG data, with ZuCo 1 containing 12 participants and ZuCo 2 18 participants \citep{Hollenstein2018zuco1, hollenstein-etal-2020-zuco}. See Table~\ref{tab:english} for more information.

In addition to these English-only datasets, there are a few multilingual datasets containing English subsets: GECO, GECO-CN, WebQMGaze, MECO L1 $1^{st}$ and $2^{nd}$ wave, Developmental reading, Grammatical/Ungrammatical, Story Reading, and Dundee. Please refer to Section \ref{sec:multilingual} for more details on these datasets.

\begin{landscape}
\begin{footnotesize}
\begin{xltabular}{1.6\textwidth}
{>{\raggedright\hsize=.1\hsize}X|>{\raggedright\hsize=.09\hsize}X>{\raggedright\hsize=.08\hsize}X>{\raggedright\hsize=.08\hsize}X>{\raggedright\hsize=.09\hsize}X>{\raggedright\hsize=.09\hsize}X>{\raggedright\hsize=.09\hsize}X>{\raggedright\hsize=.1\hsize}X>{\raggedright\hsize=.1\hsize}X>{\raggedright\hsize=.09\hsize}X>{\raggedright\hsize=.09\hsize}X}\caption{Datasets with English stimuli. For a description of the columns please refer to Table~\ref{tab:column_descriptions}.} \label{tab:english} \\
\toprule
Name & \# Participants & Age range & Age mean±SD & \# Items & Mean±SD words per item & Mean±SD chars per item & Stimulus naturalness & Stimulus length category & Eye-tracker & Sampling frequency (Hz) \\
\\\midrule
\endfirsthead
\multicolumn{11}{c}%
{\tablename\ \thetable{} -- continued from previous page} \\
\toprule
Name & \# Participants & Age range & Age mean±SD & \# Items & Mean±SD words per item & Mean±SD chars per item & Stimulus naturalness & Stimulus length category & Eye-tracker & Sampling frequency (Hz) \\
 \\\midrule
\endhead
ASD Data & 109 &  & 33.73±8.36 & 27 &  &  & Naturalistic & Text passages & Gazepoint GP3 & 60 \\
Alzheimer & 162 & 53-96 & 68.78 & 1 &  &  & Naturalistic & Text passages &  &  \\
CELER & 365 &  & 27.3±6.8 & 156 &  &  & Naturalistic & Single sentences & EyeLink 1000 Plus & 1000 \\
CFILT Coreference & 14 & 47-50 (2 expert linguists); 20-30 (12 post-graduates) &  & 22 &  &  & Naturalistic & Text passages & EyeLink 1000 Plus & 500 \\
CFILT Essay Grading & 8 &  & 25 & 48 &  &  & Naturalistic & Text passages & EyeLink 1000 & 500 \\
CFILT Sarcasm & 7 &  &  & 1000 &  &  & Naturalistic & Single sentences & EyeLink 1000 & 500 \\
CFILT Scanpath & 16 & 47-50 (3 expert linguists); 20-30 (13 post-graduates) &  & 32 &  &  & Naturalistic & Text passages & EyeLink 1000 Plus &  \\
CFILT Sentiment & 5 &  &  & 1059 &  &  & Naturalistic & Single sentences & Tobii TX 300 & 500 \\
CFILT Text Quality & 20 & 20-25 &  & 30 &  &  & Naturalistic & Text passages & EyeLink 1000 & 500 \\
CoLAGaze & 42 & 19-62 & 34.5±2 & 306 & 8.88 (gram.), 8.87 (ungram.) & 39.25 (gram.), 39.21 (ungram.) & Naturalistic & Single sentences & EyeLink Portable Duo & 2000 \\
EMTeC & 107 & 18-69 & 34.1±13.5 & 42 & 86.52±20.68 & 446.44 & Naturalistic & Text passages & EyeLink Portable Duo & 2000 \\
GGTG & 24 & 19-53 & 27.2±7.0 & 18 & 441.2±8.8 & 2442.4±297 & Naturalistic & Text passages & EyeLink 1000 Plus & 1000 \\
GazeBase - Reading Task & 322 & 18-47 & 21.89±4.22 & 18 &  &  & Naturalistic & Text passages & EyeLink 1000 & 1000 \\
GazeBase-VR & 407 & 18-58 & 20.95±4.0 & 4 &  & 820 & Naturalistic & Text passages & ET-enabled virtual-reality (VR) headset & 250 \\
IITB-HGC & 5 & 21-25 &  & 500 & 105 &  &  &  & EyeLink 1000 Plus  & 2000 \\
InteRead & 50 & 20-47 & 27.51±5.55 & 28 & 154±22.3  &  & Naturalistic & Text passages & Tobii Pro Spectrum & 1200 \\
MECO L2 $1^{st}$ Wave & 543 &  & 23.4 & 12 &  &  & Naturalistic & Text passages & EyeLink Portable Duo, EyeLink 1000, EyeLink 1000 Plus & 1000 \\
MECO L2 $2^{nd}$ Wave & 660 &  &  & 12 &  &  & Naturalistic & Text passages & EyeLink Portable Duo, EyeLink II, EyeLink 1000, EyeLink 1000 Plus  & 1,000; Serbian: 500 \\
MQA-RC & 28 &  &  & 32 &  &  & Naturalistic & Text passages & Tobii & 600 \\
Multi Choice Answering & 71 &  &  &  &  &  & Naturalistic & Text passages & Tobii TX300 screen-based & 300 \\
Not Batting an Eye & 50 &  & 29.5±3.99 & 40 &  &  & Constructed & Single sentences & EyeLink 1000 Plus & 1000 \\
OASST-ETC & 24 &  & 31.66±6.13 & 360 & 57.82±34.94 & 278.11 & Naturalistic &  & GP3 HD Eye Tracker & 60 \\
OneStop Eye Movements & 360 &  & 22.8±5.6 & 30 &  &  & Naturalistic & Text passages & EyeLink 1000 Plus & 1000 \\
Passage Reading & 48 &  & 26.48±14.83 & 40 &  &  & Naturalistic & Text passages & EyeLink 1000 & 1000 \\
Provo Corpus & 84 &  &  & 55 &  &  & Naturalistic & Text passages &  &  \\
Reading Brain & 51 & 18-40 &  & 5 & 306.2±2.8 & 1533.8±73.0 & Naturalistic & Text passages & EyeLink 1000 Plus & 1000 \\
Reading Brain L2 & 56 &  &  & 5 & 306.2±2.8 & 1533.8±73.0 & Naturalistic & Text passages &  &  \\
SB-SAT & 95 &  &  & 4 &  &  & Naturalistic & Text passages & EyeLink 1000 & 1000 \\
TECO & 41 & 18-25 & 21 & 30 & 335.07±35.88 & 1648.54 & Naturalistic & Text passages & EyeLink 1000 & 1000 \\
Task effects in human reading & 22 &  &  & 20 & 323 &  & Naturalistic & Text passages & EyeLink 1000 & 2000 \\
UCL Corpus & 43 &  & 25.8 & 205 &  &  & Naturalistic & Single sentences & EyeLink II & 500 \\
ZuCo 1 & 12 &  & 38±9.8 & 1107 &  &  & Naturalistic & Single sentences & EyeLink 1000 Plus & 500 \\
ZuCo 2 & 18 &  & 34±8.3 & 739 &  &  & Naturalistic & Single sentences & EyeLink 1000 Plus & 500 \\

\bottomrule
\end{xltabular}
\end{footnotesize}
\end{landscape}%

\subsection{Estonian}
MECO L1 $1^{st}$ wave contains an Estonian subset as specified in Section~\ref{sec:multilingual}.

\subsection{Finnish}
MECO L1 $1^{st}$ wave contains a Finnish subset as specified in Section~\ref{sec:multilingual}.

\subsection{French}
The Dundee corpus contains a French subset as specified in Section~\ref{sec:multilingual}.

\subsection{German}

The PopSci Corpus includes 17 participants reading 16 popular science texts from different sources in German \citep{popsci}. 
The Potsdam Textbook Corpus (PoTeC) contains data from 75 participants reading excerpts from either biology or physics textbooks. All participants are either beginners or experts in exactly one of the domains \citep{jakobi2025PoTeCGermanNaturalistic}.
The FakeNewsPerception Corpus includes 25 participants reading short news-style texts, each presented as a minimal pair of real and fake versions \citep{sumer2021FakeNewsPerceptionEyeMovement}. 

The Potsdam Sentence Corpus (PSC) comprises eye-tracking data from 222 participants reading isolated German sentences, which are constructed around a target word \citep{Hohenstein2017PSC, kliegl2004psc}. There exist various other datasets where the same stimuli have been used:
PSC II (Potsdam Sentence Corpus II) contains eye‑tracking data from 149 participants on the same texts \citep{heister2012AnalysingLargeDatasets}.
In addition, \cite{dimigen2011CoregistrationEyeMovements} co‑registered eye‑tracking and EEG data from 30 participants reading the PSC sentences.
In the Eye‑Voice Span study, 63 participants' eye movements and voice were recorded (half of the participants reading aloud, the other half silently) to investigate the temporal distance between eyes and voice during reading aloud, again using the PSC sentences \citep{laubrock2015EyevoiceSpanReading}.
The Parafoveal‑on‑Foveal study includes 111 participants (several age groups) reading the PSC sentences to examine parafoveal‑on‑foveal effects \citep{wotschack2013ReadingStrategyModulates}.
See Table~\ref{tab:german} for more detailed information on German datasets.

Four multilingual datasets contain one or multiple German subsets: WebQAmGaze, Grammatical/Ungrammatical, and  MECO L1 $1^{st}$ and $2^{nd}$ wave (for details refer to Section~\ref{sec:multilingual}).

\begin{landscape}
\begin{footnotesize}
\begin{xltabular}{1.6\textwidth}
{>{\raggedright\hsize=.1\hsize}X|>{\raggedright\hsize=.09\hsize}X>{\raggedright\hsize=.08\hsize}X>{\raggedright\hsize=.08\hsize}X>{\raggedright\hsize=.09\hsize}X>{\raggedright\hsize=.09\hsize}X>{\raggedright\hsize=.09\hsize}X>{\raggedright\hsize=.1\hsize}X>{\raggedright\hsize=.1\hsize}X>{\raggedright\hsize=.09\hsize}X>{\raggedright\hsize=.09\hsize}X}\caption{Datasets with German stimuli. For a description of the columns please refer to Table~\ref{tab:column_descriptions}.} \label{tab:german} \\
\toprule
Name & \# Participants & Age range & Age mean±SD & \# Items & Mean±SD words per item & Mean±SD chars per item & Stimulus naturalness & Stimulus length category & Eye-tracker & Sampling frequency (Hz) \\
\midrule
\endfirsthead
\multicolumn{11}{c}%
{\tablename\ \thetable{} -- continued from previous page} \\
\toprule
Name & \# Participants & Age range & Age mean±SD & \# Items & Mean±SD words per item & Mean±SD chars per item & Stimulus naturalness & Stimulus length category & Eye-tracker & Sampling frequency (Hz) \\
\midrule
\endhead
Eye-voice span & 63 & 16-24 & 18.6 & 144 & 8.54±1.44 & 54.58±10.67 & Partially constructed & Single sentences & EyeLink 1000 &  \\
Fake-News-Perception & 25 &  & 25.9±4.8 & 120 &  &  & Naturalistic & Text passages & Tobii Pro Spectrum & 600 \\
PSC & 222 & 16-84 &  & 144 &  &  & Partially constructed & Single sentences & EyeLink I, EyeLink II & 250 / 500 \\
PSC II & 273 &  &  & 144 & 8.54±1.44 & 54.58±10.67 & Partially constructed & Single sentences &  &  \\
PSC with EEG & 30 & 17-37 & 23 & 144 & 7.9 &  & Partially constructed & Single sentences & IView-X & 240 \\
Parafoveal-on-foveal & 111 & 16-76 &  & 144 &  &  & Partially constructed & Single sentences & EyeLink II, EyeLink I (old adults easy question group) & 500, 250 (old adults, easy question group) \\
PoTeC & 75 &  & 24.2±4.2 & 12 &  &  & Naturalistic & Text passages & EyeLink 1000 & 1000 \\
PopSci Corpus & 17 &  &  & 16 &  &  & Naturalistic & Text passages & EyeLink 1000 & 1000 \\
\bottomrule
\end{xltabular}
\end{footnotesize}
\end{landscape}%

\subsection{Greek}
MECO L1 $1^{st}$ wave contains a Greek subset as specified in Section~\ref{sec:multilingual}.

\subsection{Hebrew}
MECO L1 $1^{st}$ wave contains a Hebrew subset as specified in Section~\ref{sec:multilingual}.

\subsection{Hindi}
The Potsdam-Allahabad Hindi Eyetracking Corpus and MECO L1 $2^{nd}$ wave contain a Hindi subset as specified in Section \ref{sec:multilingual}.

\subsection{Icelandic}
MECO L1 $2^{nd}$ wave contains an Icelandic subset as specified in Section \ref{sec:multilingual}.

\subsection{Italian}
MECO L1 $1^{st}$ wave contains an Italian subset as specified in Section~\ref{sec:multilingual}.

\subsection{Japanese}

One monolingual dataset exists for Japanese. It contains 24 participants reading newspaper articles from \textit{The Balanced Corpus of Contemporary Written Japanese} \citep{BCCWJ-eye-tracking}. Please note that the information available for this dataset is limited which is why there is no dedicated table for Japanese but it is included in the online overview.

The Story Reading dataset contains a Japanese subset, please refer to Section~\ref{sec:multilingual}.

\subsection{Korean}
MECO L1 $1^{st}$ wave and the Story Reading corpus contain a Korean subset as specified in Section~\ref{sec:multilingual}.

\subsection{Mongolian}

For Mongolian, a separate MECO L1 dataset has been collected comprising 66 participants reading 12 Wikipedia-style texts as in MECO \citep{baoEyeMovementDatabase2025}. See Table~\ref{tab:mongolian} for more information.

%\begin{landscape}
\begin{footnotesize}
\begin{xltabular}{\textwidth}
{X|XXXXXXX}\caption{Datasets with Mongolian stimuli. For a description of the columns please refer to Table~\ref{tab:column_descriptions}.} \label{tab:mongolian} \\
\toprule
Name & \# Participants & Age range &  \# Items & Stimulus naturalness & Stimulus length category & Eye-tracker & Sampling frequency (Hz) \\
\midrule
\endfirsthead
\multicolumn{8}{c}%
{\tablename\ \thetable{} -- continued from previous page} \\
\toprule
Name & \# Participants & Age range &  \# Items & Stimulus naturalness & Stimulus length category & Eye-tracker & Sampling frequency (Hz) \\
\midrule
\endhead
MECO Mongolian & 66 & 18-27 & 12 & Naturalistic & Text passages & EyeLink 1000 & 1000 \\
\bottomrule
\end{xltabular}
\end{footnotesize}
%\end{landscape}%

\subsection{Norwegian}
The Irony Processing dataset includes reading data from 52 adults diagnosed with ADHD and 55 neurotypical participants reading 24 stories that included either literal or ironic statements \citep{kyriacouIronyProcessingAdults2025}.
See Table~\ref{tab:norwegian} for more information.

MECO L1 $1^{st}$ and $2^{nd}$ waves contain a Norwegian subset, please refer to Section~\ref{sec:multilingual}.

%\begin{landscape}
\begin{footnotesize}
\begin{xltabular}{\textwidth}
{X|XXXXXX}\caption{Datasets with Norwegian stimuli. For a description of the columns please refer to Table~\ref{tab:column_descriptions}.} \label{tab:norwegian} \\
\toprule
Name & \# Participants & Age range & Age mean±SD & Stimulus type & Eye-tracker & Sampling frequency (Hz) \\\midrule
\endfirsthead
\multicolumn{7}{c}%
{\tablename\ \thetable{} -- continued from previous page} \\
\toprule
Name & \# Participants & Age range & Age mean±SD & Stimulus type & Eye-tracker & Sampling frequency (Hz) \\\midrule
\endhead
Irony Processing & 107 & 18-35 & 25.40 & constructed text passages & EyeLink 1000 plus & 1000 \\
\bottomrule
\end{xltabular}
\end{footnotesize}

\subsection{Persian}

The PSR (Persian Sentence Corpus; \citealp{tekbudak2024PSRCorpusPersian}) is a Persian dataset containing data from 60 participants reading 99 single sentences. See Table~\ref{tab:persian} for more information.

%\begin{landscape}
\begin{footnotesize}
\begin{xltabular}{\textwidth}
{>{\raggedright\hsize=.07\hsize}X|>{\raggedright\hsize=.09\hsize}X>{\raggedright\hsize=.12\hsize}X>{\raggedright\hsize=.1\hsize}X>{\raggedright\hsize=.13\hsize}X>{\raggedright\hsize=.14\hsize}X>{\raggedright\hsize=.15\hsize}X>{\raggedright\hsize=.1\hsize}X>{\raggedright\hsize=.1\hsize}X}\caption{Datasets with Persian stimuli. For a description of the columns please refer to Table~\ref{tab:column_descriptions}.} \label{tab:persian} \\
\toprule
Name & \# Participants & Age mean±SD & \# Items & Mean±SD words per item & Stimulus naturalness & Stimulus length category & Eye-tracker & Sampling frequency (Hz) \\
\midrule
\endfirsthead
\multicolumn{9}{c}%
{\tablename\ \thetable{} -- continued from previous page} \\
\toprule
Name & \# Participants & Age mean±SD & \# Items & Mean±SD words per item & Stimulus naturalness & Stimulus length category & Eye-tracker & Sampling frequency (Hz) \\
\midrule
\endhead
PSR & 60 & 29.60±4.03 & 99 & 10.27±1.01 & Naturalistic & Single sentences & EyeLink 1000 Plus & 1000 \\
\bottomrule
\end{xltabular}
\end{footnotesize}
%\end{landscape}%

\subsection{Portuguese}
RastrOS is a Portuguese dataset including 37 participants reading 50 naturalistic text passages from various sources \citep{Leal2022rastros}. See Table~\ref{tab:portuguese} for more information.  

The MECO L1 $2^{nd}$ wave contains a Brazilian Portuguese subset as specified in Section \ref{sec:multilingual}.

%\begin{landscape}
\begin{footnotesize}
\begin{xltabular}{\textwidth}
{>{\raggedright\hsize=.1\hsize}X|>{\raggedright\hsize=.15\hsize}X>{\raggedright\hsize=.15\hsize}X>{\raggedright\hsize=.09\hsize}X>{\raggedright\hsize=.13\hsize}X>{\raggedright\hsize=.13\hsize}X>{\raggedright\hsize=.1\hsize}X>{\raggedright\hsize=.15\hsize}X}\caption{Datasets with Portuguese stimuli. For a description of the columns please refer to Table~\ref{tab:column_descriptions}.} \label{tab:portuguese} \\
\toprule
Name & \# Participants & Age mean±SD & \# Items & Stimulus naturalness & Stimulus length category & Eye-tracker & Sampling frequency (Hz) \\
\midrule
\endfirsthead
\multicolumn{8}{c}%
{\tablename\ \thetable{} -- continued from previous page} \\
\toprule
Name & \# Participants & Age mean±SD & \# Items & Stimulus naturalness & Stimulus length category & Eye-tracker & Sampling frequency (Hz) \\
\midrule
\endhead
RastrOS & 37 & 22.2±4.7 & 50 & Naturalistic & Text passages & EyeLink 1000 & 1000 \\
\bottomrule
\end{xltabular}
\end{footnotesize}
%\end{landscape}%

\subsection{Russian}
The RSC (Russian Sentence Corpus) includes data from 96 participants. The participants read 144 sentences that were constructed around a target word \citep{Laurinavichyute2018RSC}.
See Table~\ref{tab:russian} for more information.

MECO L1 $1^{st}$ and $2^{nd}$ waves contain a Russian subset as specified in Section~\ref{sec:multilingual}. 

%\begin{landscape}
\begin{footnotesize}
\begin{xltabular}{\textwidth}
{>{\raggedright\hsize=.08\hsize}X|>{\raggedright\hsize=.15\hsize}X>{\raggedright\hsize=.15\hsize}X>{\raggedright\hsize=.1\hsize}X>{\raggedright\hsize=.12\hsize}X>{\raggedright\hsize=.15\hsize}X>{\raggedright\hsize=.1\hsize}X>{\raggedright\hsize=.15\hsize}X}\caption{Datasets with Russian stimuli. For a description of the columns please refer to Table~\ref{tab:column_descriptions}.} \label{tab:russian} \\
\toprule
Name & \# Participants & Age mean±SD & \# Items & Stimulus naturalness & Stimulus length category & Eye-tracker & Sampling frequency (Hz) \\
\midrule
\endfirsthead
\multicolumn{8}{c}%
{\tablename\ \thetable{} -- continued from previous page} \\
\toprule
Name & \# Participants & Age mean±SD & \# Items & Stimulus naturalness & Stimulus length category & Eye-tracker & Sampling frequency (Hz) \\
\midrule
\endhead
RSC & 96 & 24.0 & 144 & Naturalistic & Single sentences & EyeLink 1000 Plus & 1000 \\
\bottomrule
\end{xltabular}
\end{footnotesize}
%\end{landscape}%

\subsection{Serbian}
For one of two related Serbian datasets, eye-tracking and EEG data of in total 25 $2^{nd}$ and $3^{rd}$ grade children was collected \citep{jakovljevic2021SensorHubDetecting}. The children read one story split into multiple pages.
For the same text, data from dyslexic children was recorded as well \citep{jakovljevic2021RelationPhysiologicalParameters}.
For more information see Table~\ref{tab:serbian}.

MECO L1 $2^{nd}$ wave contains a Serbian subset as specified in Section~\ref{sec:multilingual}.

%\begin{landscape}
\begin{footnotesize}
\begin{xltabular}{\textwidth}
{>{\raggedright\hsize=.11\hsize}X|>{\raggedright\hsize=.15\hsize}X>{\raggedright\hsize=.15\hsize}X>{\raggedright\hsize=.1\hsize}X>{\raggedright\hsize=.12\hsize}X>{\raggedright\hsize=.15\hsize}X>{\raggedright\hsize=.1\hsize}X>{\raggedright\hsize=.12\hsize}X}\caption{Datasets with Serbian stimuli. For a description of the columns please refer to Table~\ref{tab:column_descriptions}.} \label{tab:serbian} \\
\toprule
Name & \# Participants & Age range & \# Items & Stimulus naturalness & Stimulus length category & Eye-tracker & Sampling frequency (Hz) \\
\midrule
\endfirsthead
\multicolumn{8}{c}%
{\tablename\ \thetable{} -- continued from previous page} \\
\toprule
Name & \# Participants & Age range & \# Items & Stimulus naturalness & Stimulus length category & Eye-tracker & Sampling frequency (Hz) \\
\midrule
\endhead
Children EEG & 25 & 8-9 & 1 & Naturalistic & Text passages & SMI RED-m & 120 \\
Dyslexic children EEG & 36 & 8-12 & 1 & Naturalistic & Text passages & SMI RED-m & 120 \\
\bottomrule
\end{xltabular}
\end{footnotesize}
%\end{landscape}%

\subsection{Spanish}
Cuentos is a Spanish dataset containing data from 113 native Spanish adults \citep{travi2026CuentosLargeScaleEyeTracking}. The participants read both long and short stories. For more information see Table~\ref{tab:spanish}.

MECO L1 $1^{st}$ and $2^{nd}$ waves and WebQAmGaze contain a Spanish subset as specified in Section~\ref{sec:multilingual}.

\begin{footnotesize}
\begin{xltabular}{\textwidth}
{X|XXXXXXX}\caption{Datasets with Spanish stimuli. For a description of the columns please refer to Table~\ref{tab:column_descriptions}.} \label{tab:spanish} \\
\toprule
Name & \# Participants &  Age mean±SD & \# Items & Mean±SD words per item & Stimulus naturalness &   Eye-tracker & Sampling frequency (Hz) \\
\midrule
\endfirsthead
\multicolumn{8}{c}%
{\tablename\ \thetable{} -- continued from previous page} \\
\toprule
Name & \# Participants & Age mean±SD & \# Items & Mean±SD words per item & Stimulus naturalness &  Eye-tracker & Sampling frequency (Hz) \\
\midrule
\endhead
Cuentos & 113 & 23.8±7.78 & 30 & 3300±747 (long), 795±135 (short) & Naturalistic &  EyeLink 1000 & 1000 \\
\bottomrule
\end{xltabular}
\end{footnotesize}

\subsection{Swedish}

The Swedish Dyslexia datasets contains data from 185 children reading a text \citep{benfattoScreeningDyslexiaUsing2016}. The children are either at high or low risk of being affected by dyslexia. See Table~\ref{tab:swedish} for more information.

%\begin{landscape}
\begin{footnotesize}
\begin{xltabular}{\textwidth}
{X|XXXXXX}\caption{Datasets with Swedish stimuli. For a description of the columns please refer to Table~\ref{tab:column_descriptions}.} \label{tab:swedish} \\
\toprule
Name & \# Participants & Age range & \# Items & Mean±SD words per item & Eye-tracker & Sampling frequency (Hz) \\
\midrule
\endfirsthead
\multicolumn{7}{c}%
{\tablename\ \thetable{} -- continued from previous page} \\
\toprule
Name & \# Participants & Age range & \# Items & Mean±SD words per item & Eye-tracker & Sampling frequency (Hz) \\
\midrule
\endhead
Swedish Dyslexia & 185 & 9-10 & 1 & 46 & Ober-2TM & 100 \\
\bottomrule
\end{xltabular}
\end{footnotesize}
%\end{landscape}%

\subsection{Turkish}

TURead is a Turkish dataset that contains data from 196 participants reading 192 stimulus sentences of which few span more than one sentence. The sentences are constructed around different target words \citep{acarturk2024TUReadEyeMovement}. See Table~\ref{tab:turkish} for more information.

Furthermore, MECO L1 $1^{st}$ and $2^{nd}$ waves contain a Turkish subset as specified in Section~\ref{sec:multilingual}. 

%\begin{landscape}
\begin{footnotesize}
\begin{xltabular}{\textwidth}
{>{\raggedright\hsize=.11\hsize}X|>{\raggedright\hsize=.15\hsize}X>{\raggedright\hsize=.15\hsize}X>{\raggedright\hsize=.1\hsize}X>{\raggedright\hsize=.12\hsize}X>{\raggedright\hsize=.15\hsize}X>{\raggedright\hsize=.1\hsize}X>{\raggedright\hsize=.12\hsize}X}\caption{Datasets with Turkish stimuli. For a description of the columns please refer to Table~\ref{tab:column_descriptions}.} \label{tab:turkish} \\
\toprule
Name & \# Participants & Age mean±SD & \# Items & Stimulus naturalness & Stimulus length category & Eye-tracker & Sampling frequency (Hz) \\
\midrule
\endfirsthead
\multicolumn{8}{c}%
{\tablename\ \thetable{} -- continued from previous page} \\
\toprule
Name & \# Participants & Age mean±SD & \# Items & Stimulus naturalness & Stimulus length category & Eye-tracker & Sampling frequency (Hz) \\
\midrule
\endhead
TURead & 196 & 22.7±2.6 & 192 & Naturalistic & Both single sentences and text passages & EyeLink 1000 & 1000 \\
\bottomrule
\end{xltabular}
\end{footnotesize}
%\end{landscape}%

\subsection{Uighur}
Morphological Structure I and Morphological Structure II contain data for the Uighur language as specified in Section~\ref{sec:multilingual}.

\subsection{Urdu}
The Potsdam-Allahabad Hindi Eyetracking Corpus contains an Urdu subset as specified in Section \ref{sec:multilingual}.

\section{The \pymovements dataset library}
\label{sec:pm}
In order to increase the re-usability of the datasets presented in this review, we have integrated all publicly available datasets into the open-source Python package \pymovements which is a library to preprocess eye-tracking data including an eye-tracking dataset library \citep[][for more information see below]{Krakowczyk_pymovementsETRA2023}. Integrating a dataset means that the dataset is made available as a resource within the package. A description of the dataset is added including links pointing to the respective resources and references of the original publication(s). \pymovements allows users to download all integrated datasets for their own use and, optionally, process the data directly using the package (\citealp{pm-datasets-2025}; please refer to Listing~\ref{lst:download} for a code example on how to download a dataset). In order to do so, the published recording of a dataset is parsed into a standardized format \textit{without changing the recorded data or discarding any of the data}. To familiarize yourself with \pymovements' data structures please consult their \href{https://github.com/pymovements/pymovements}{website} and \href{https://pymovements.readthedocs.io/en/stable/tutorials/index.html}{tutorials}.

\begin{lstlisting}[language=Python,caption={Example of how to download a dataset in the Python library \pymovements. For more information please refer to their \href{https://pymovements.readthedocs.io/en/stable/tutorials/index.html}{tutorials} as those are updated regularly. The list of the integrated datasets can be found on \pymovements' \href{https://pymovements.readthedocs.io/en/stable/datasets/index.html}{website} as well in the living online table.},captionpos=t,label=lst:download]
import pymovements as pm

# Specify the dataset name and the local data directory.
dataset = pm.Dataset(name='PoTeC', path='data/')

# Download the dataset.
dataset.download()

\end{lstlisting}

If a dataset is published with precomputed gaze events or reading measures, these can also be integrated and downloaded.\footnote{Please consult this tutorial for instructions on how to download public datasets: \url{https://pymovements.readthedocs.io/en/stable/tutorials/public-datasets.html}} It is important to note that \pymovements \textit{does not host the datasets}; rather, it serves as an interface that facilitates working with datasets published \textit{by their original authors}. Users are always required to cite the corresponding dataset publication when using data accessed through \pymovements. Not all datasets can be integrated into \pymovements, as integration depends on the availability of the underlying data. A list of integrated datasets is provided on the package's \href{https://pymovements.readthedocs.io/en/stable/datasets/index.html}{documentation website}. Instructions on how to add new datasets to \pymovements are described in Section~\ref{sec:add}.

\texttt{pymovements'} core functionalities include parsing original eye-tracker data files produced by various manufacturers, extracting and formatting gaze samples and metadata into a device-independent standardized data frame, and providing preprocessing methods. The package also supports gaze event detection using a selection of widely adopted, pre-implemented algorithms (I-VT~\citealp{salvucci2000fixations}, I-DT~\citealp{salvucci2000fixations}, microsaccades~\citealp{engbert2003MicrosaccadesUncoverOrientation}), the computation of event properties such as saccade amplitude and fixation dispersion, and several visualization techniques (e.g., main sequence plots). All of \pymovements' functions can be applied to custom datasets stored locally on the user's device as well as to datasets listed in the package's overview. Consequently, any dataset can be efficiently preprocessed into all formats described in Section~\ref{sec:data-formats} using the \textit{same standardized pipeline}. Again, the underlying data will not be changed as the standardization refers only to the data formats, and the implementation of the algorithms. 

Although the preprocessing pipeline is standardized, users retain full flexibility to select parameters and customize the pipeline to suit their specific needs. Importantly, when two users apply the same algorithm, the implementation is identical, ensuring consistency and eliminating discrepancies caused by differing implementations. Furthermore, since \pymovements employs a rigorous testing scheme for all functionalities, implementation errors are minimized, thereby substantially enhancing the overall quality of preprocessing.

\section{Adding and editing datasets}
\label{sec:add}
As the datasets in this review are integrated into the Python library \pymovements as well as the filterable online table, it is possible to add new datasets to both channels as well as submit a request to edit existing datasets. In order to add or edit datasets in the online table, dataset creators are asked to send a request through our website. Dataset creators have to complete a form where they fill in all information on their dataset that is available. More information can be found at the following link: \href{https://t.uzh.ch/1Yv}{Add your dataset}.

In order to add datasets to \pymovements or edit already added ones, dataset authors are asked to create an issue on the package's GitHub page using a template specifically designed for adding new datasets: \href{https://github.com/aeye-lab/pymovements/issues/new?template=DATASET.md}{dataset issue template}. Alternatively, an email can be sent.\footnote{\href{mailto:pymovements@python.com}{pymovements@python.com}} The \pymovements maintenance team is happy to provide assistance through the process to enable fast and easy dataset integration. Each integrated dataset is tested by applying standard preprocessing steps to the resource.

\section{Discussion \& future work}
Eye-tracking-while-reading data is widely used across different disciplines. While the increase in the number of datasets and their overall growth in size is substantial, the data is oftentimes hard to find and rarely fully reusable as it is often not made available in different formats or is lacking  necessary metadata. In case the data is made publicly available, it is often highly aggregated, via discipline-specific platforms, and not interoperable. This is caused by a lack of standards in the field on how to document, share and publish eye-tracking-while-reading data.  
Publicly available datasets including comprehensive metadata are needed to, for example, verify research outcomes by reproducing analyses or replicating experiments. The recent interest of researchers in using machine-learning methods to work with eye-tracking-while-reading data for various use cases, including data-driven generative modeling of eye movements, further intensifies the need for huge amounts of publicly available data.

In sum, the eye-tracking-while-reading community is in need for standards that enable researchers to implement the FAIR principles, and common sharing practices across different disciplines. Recent data collection projects such as MECO\footnote{\url{https://meco-read.com/}} \citep{Siegelman2022mecol1wave1, siegelman2025mecol1wave2, kuperman-2023-mecol2-wave1, kuperman2025mecol2wave2} and MultiplEYE\footnote{\url{https://multipleye.eu/}} explicitly emphasize the public and complete sharing of their data, with the latter publishing the entire preprocessing pipeline as well as the data in all formats including extensive metadata \citep{Jakobi2025MultipleyeETRA}. 
Our goal with this eye-tracking-while-reading dataset review is to explicitly strengthen such existing endeavors and promote the FAIR principles within eye-tracking-while-reading research. By presenting our overview in an easily accessible filterable online table (\url{https://t.uzh.ch/1Yh}), we intend to make the data more \textit{findable}. Integrating the data into the dataset library of the open-source Python package \pymovements will make the datasets more \textit{accessible} as well as \textit{interoperable}. Overall, the data will become easier to \textit{reuse}. Last but not least, the living nature of the overview should encourage authors to publish their data in full and add it to the overview, supporting the adoption of FAIR principles for future datasets as well.

\section*{Acknowledgements}
 We thank Damian Hiltebrand for creating a first draft of the published online table, as well as Shan Gao for assisting with formatting of the table.

\section*{Open practices statement}
The living online table presented in this paper can be openly accessed by anyone. As no data has been collected for this work, there are no further statements.

\newpage

\section*{Declarations}

\subsection*{Funding}
This work was partially funded by the Swiss National Science Foundation under grant 212276 (MeRID) and grant 220330 (EyeNLG), and is supported by COST Action MultiplEYE, CA21131.
\subsection*{Competing interests}
\textit{Not applicable.}
\subsection*{Ethics approval}
\textit{Not applicable.}
\subsection*{Consent to participate}
\textit{Not applicable.}
\subsection*{Consent for publication}
\textit{Not applicable.}
\subsection*{Availability of data and materials}
\textit{Not applicable.}
\subsection*{Code availability}
\textit{Not applicable.}
\subsection*{Author contribution}
\textit{Not applicable.}

%\begin{appendices}

%\section{Section title of first appendix}\label{secA1}




%\end{appendices}


\bibliography{references}% common bib file

\end{document}